\documentclass[10pt,journal,compsoc]{IEEEtran}

\ifCLASSINFOpdf
\usepackage[pdftex]{graphicx}
\graphicspath{{./figures/}}

\else

\fi

\ifCLASSOPTIONcompsoc
\usepackage[nocompress]{cite}
\else
\usepackage{cite}
\fi

\usepackage[linesnumbered,lined,noend,ruled]{algorithm2e}
\usepackage{amsmath,amssymb}
\usepackage{subfigure}
\usepackage{amstext}
\usepackage{amsfonts}
\usepackage{url}
\usepackage{bbm}
\usepackage{color}

\usepackage{tablefootnote}
\usepackage{float}

\usepackage{color}
\definecolor{red}{RGB}{255,0,0}

\usepackage{multirow}
\usepackage[export]{adjustbox}
\usepackage{ragged2e}

\usepackage{xspace}
\newcommand{\ie}{\textit{i.e.}\xspace}
\newcommand{\eg}{\textit{e.g.}\xspace}
\newcommand{\etal}{\textit{et al.}\xspace}

\usepackage[pagebackref=true,breaklinks=true,letterpaper=true,colorlinks,bookmarks=false]{hyperref}

\begin{document}
\title{Arbitrary Shape Text Detection via Segmentation with Probability Maps}
%
%

\author{Shi-Xue Zhang, Xiaobin Zhu, Lei Chen, Jie-Bo Hou, Xu-Cheng Yin,~\IEEEmembership{Senior Member,~IEEE}
\IEEEcompsocitemizethanks{
	\IEEEcompsocthanksitem Corresponding authors:Xiaobin Zhu	.
	\IEEEcompsocthanksitem S. Zhang, L. Chen, J. Hou, and X. Zhu are with the School of Computer and Communication Engineering, University of Science and Technology Beijing (USTB), Beijing, 100083, China.\protect\\
	E-mail: \{zhangshixue111,chenleiustb\}@163.com; houjiebo@gmail.com; zhuxiaobin@ustb.edu.cn.
	\IEEEcompsocthanksitem X. Yin  is with the School of Computer and Communication Engineering, and Institute of Artificial Intelligence, University of Science and Technology Beijing (USTB), Beijing, 100083, China, also with USTB-EEasyTech Joint Lab of Artificial Intelligence, Beijing, 100083, China.\protect\\
	E-mail: xuchengyin@ustb.edu.cn.
}
\thanks{Manuscript received 16 July 2020; revised 16 Mar. 2022; accepted 15 May 2022. Digital Object Identifier no. 10.1109/TPAMI.2022.3176122}
}

\markboth{Journal of \LaTeX\ Class Files,~Vol.~14, No.~8, May~2020}%
{Shell \MakeLowercase{\textit{et al.}}: Bare Demo of IEEEtran.cls for Computer Society Journals}

\IEEEtitleabstractindextext{%

\begin{abstract}
\justifying
Arbitrary shape text detection is a challenging task due to the significantly varied sizes and aspect ratios, arbitrary orientations or shapes, inaccurate annotations, etc. Due to the scalability of pixel-level prediction, segmentation-based methods can adapt to various shape texts and hence attracted considerable attention recently. However, accurate pixel-level annotations of texts are formidable, and the existing datasets for scene text detection only provide coarse-grained boundary annotations. Consequently, numerous misclassified text pixels or background pixels inside annotations always exist, degrading the performance of segmentation-based text detection methods. Generally speaking, whether a pixel belongs to text or not is highly related to the distance with the adjacent annotation boundary. With this observation, in this paper, we propose an innovative and robust segmentation-based detection method via probability maps for accurately detecting text instances. To be concrete, we adopt a Sigmoid Alpha Function (SAF) to transfer the distances between boundaries and their inside pixels to a probability map. However, one probability map can not cover complex probability distributions well because of the uncertainty of coarse-grained text boundary annotations. Therefore, we adopt a group of probability maps computed by a series of Sigmoid Alpha Functions to describe the possible probability distributions. In addition, we propose an iterative model to learn to predict and assimilate probability maps for providing enough information to reconstruct text instances. Finally, simple region growth algorithms are adopted to aggregate probability maps to complete text instances. Experimental results demonstrate that our method achieves state-of-the-art performance in terms of detection accuracy on several benchmarks.
Notably, our method with Watershed Algorithm as post-processing achieves the best F-measure on Total-Text (88.79\%), CTW1500 (85.75\%), and MSRA-TD500 (88.93\%). Besides, our method achieves promising performance on multi-oriented datasets (ICDAR2015) and multilingual datasets (ICDAR2017-MLT). Code is available at: \url{https://github.com/GXYM/TextPMs}.
	
\end{abstract}


}

\maketitle

\IEEEdisplaynontitleabstractindextext

%
\IEEEpeerreviewmaketitle

\section{Introduction}

\begin{figure}[htbp]
	\centering
	\subfigure[Object Box]{
		\begin{minipage}[t]{0.48\linewidth}
			\centering
			\includegraphics[width=4.25cm,height=3cm]{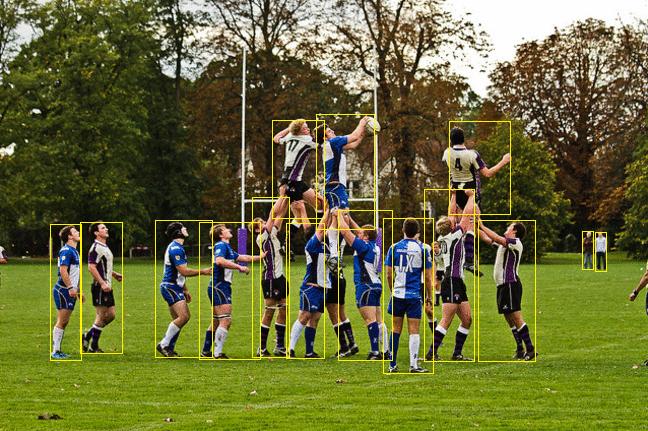}
		\end{minipage}
	}%
	\subfigure[Object Mask]{
		\begin{minipage}[t]{0.48\linewidth}
			\centering
			\includegraphics[width=4.25cm,height=3cm]{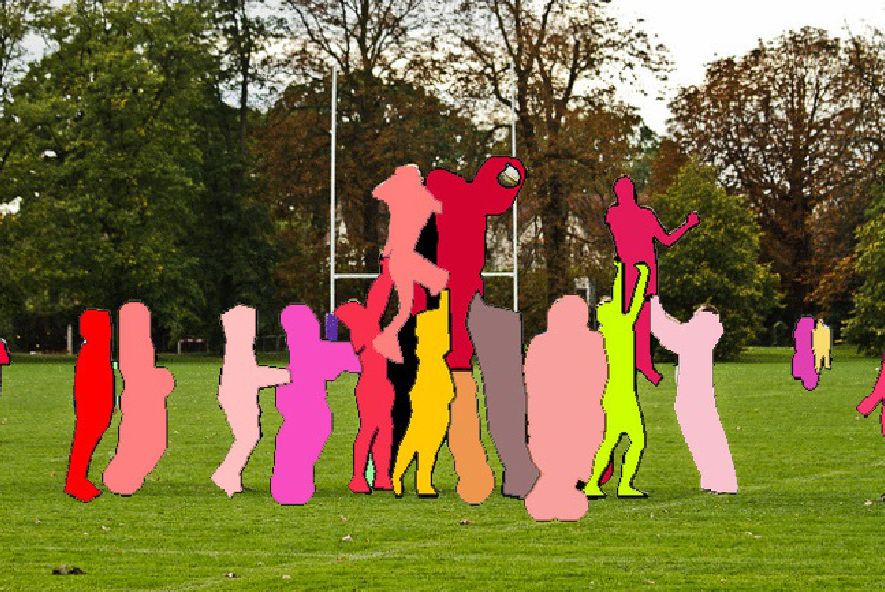}
		\end{minipage}
	}%
	\quad
	\subfigure[Text Boundary]{
		\begin{minipage}[t]{0.48\linewidth}
			\centering
			\includegraphics[width=4.3cm,height=3.3cm]{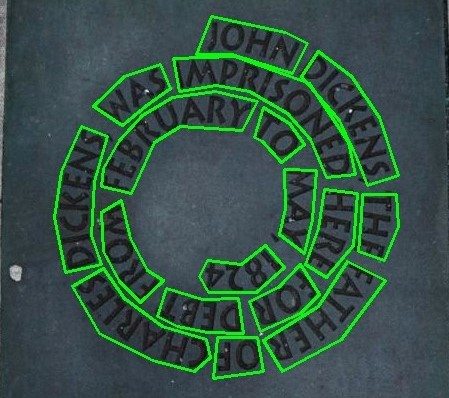}
		\end{minipage}
	}%
	\subfigure[Text Mask]{
		\begin{minipage}[t]{0.48\linewidth}
			\centering
			\includegraphics[width=4.3cm,height=3.3cm]{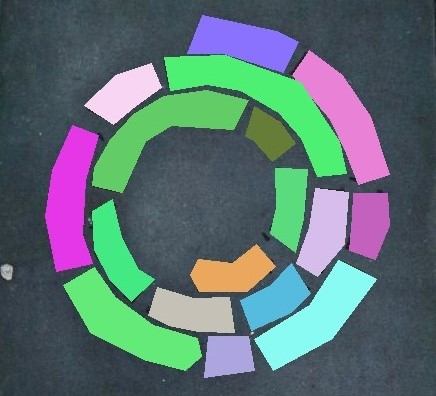}
		\end{minipage}
	}%
	\centering
	\caption{Differences of annotations: (a) box annotations for object detection, (b) pixel-level annotations for instance segmentation, (c) polygon annotations for curved text detection. (c) text instance masks obtained by polygon annotations, and (d) text masks generated by polygon annotations. General instance segmentation datasets usually provide two annotations: object box and pixel-level mask. However, scene text detection datasets only provide coarse-grained annotations. }
	\label{fig:fg1}
\end{figure}

\begin{figure*}[ht]
	\begin{center}
		\subfigure[Total-Text]{
			\begin{minipage}[t]{0.24\linewidth}
				\centering
				\includegraphics[width=4.3cm,height=3cm]{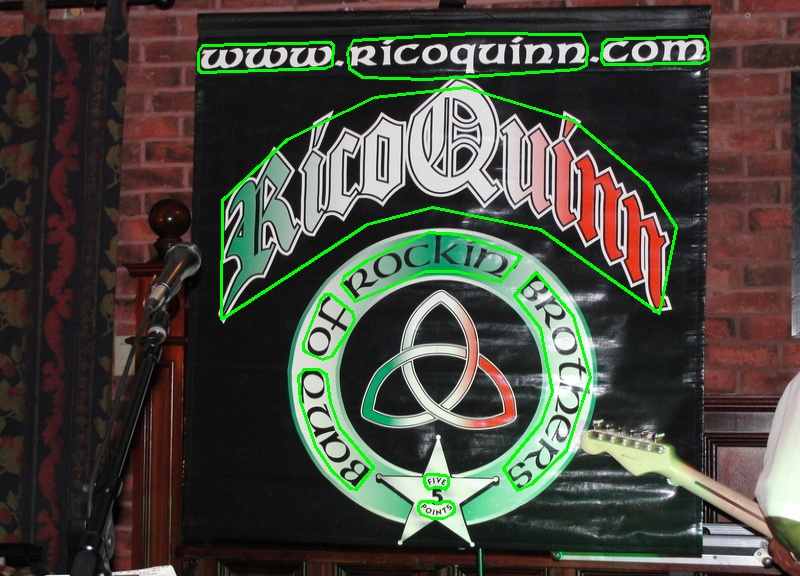}
			\end{minipage}
		}%
		\subfigure[CTW1500]{
			\begin{minipage}[t]{0.24\linewidth}
				\centering
				\includegraphics[width=4.3cm,height=3cm]{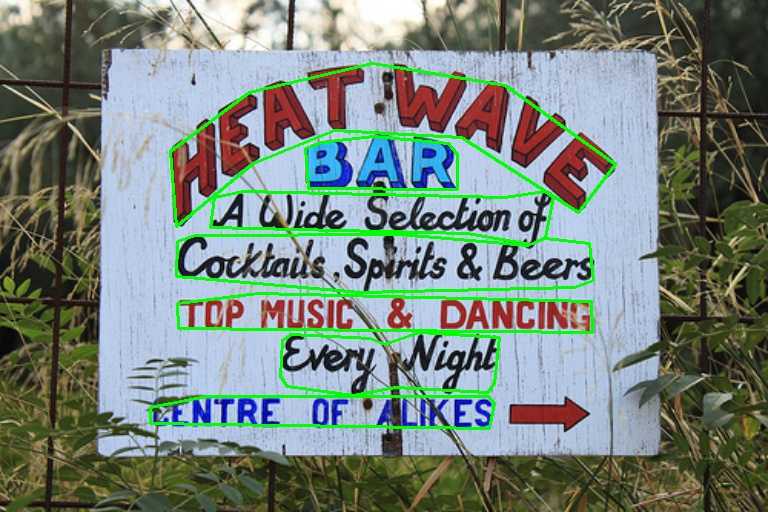}
			\end{minipage}
		}%
		\subfigure[MSRA-TD500]{
			\begin{minipage}[t]{0.24\linewidth}
				\centering
				\includegraphics[width=4.4cm,height=3cm]{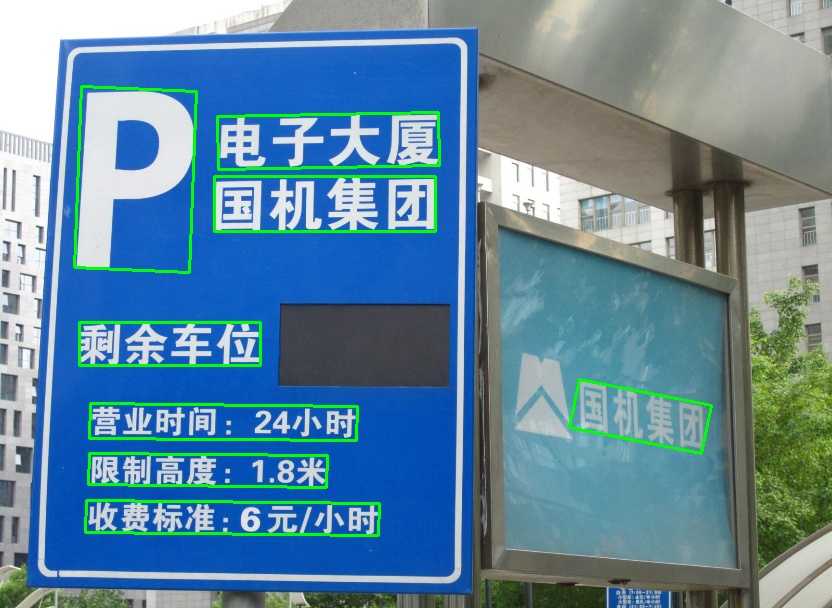}
			\end{minipage}
		}%
		\subfigure[ICDAR2017-MLT]{
			\begin{minipage}[t]{0.24\linewidth}
				\centering
				\includegraphics[width=4.4cm,height=3cm]{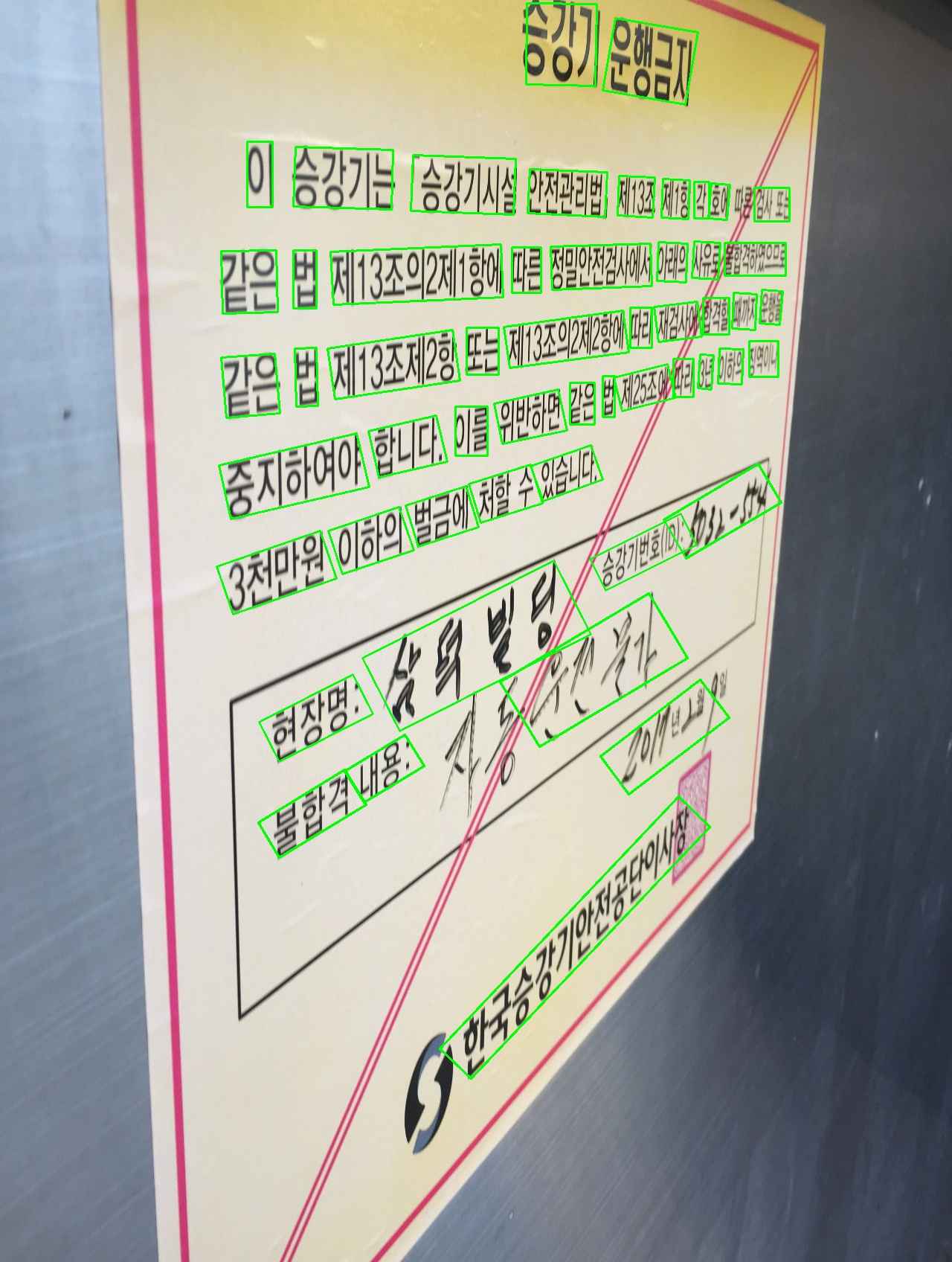}
			\end{minipage}
		}%
		\caption{Representative  detection results (enclosed by green contours) of arbitrary texts on challenging scene images from public datasets.}
		\label{fig:fg2}
	\end{center}
	\vspace{-2.0em}
\end{figure*}

\IEEEPARstart{S}{cene} text detection is a fundamental and important task, which can facilitate the development of numerous applications, such as scene parsing~\cite{Yin-Z,KangKY17,Yin-V}, product search~\cite{product_search}, and autonomous driving~\cite{drive2, drive1}. With the prosperity of deep learning, many text detection methods~\cite{CRAFT, DB, PSENet_v2, Boundary, DRRG} have been proposed and achieved impressive performance. However, scene text detection is still complicated due to arbitrary irregular rotated shapes (\eg curved form and trapezoid), varied aspect ratios, and inaccurate annotations.

Recently, Ch’ng \etal~\cite{TotalText} and Liu \etal~\cite{CTW1500} published two curved-text focused datasets, named Total-Text and CTW1500, respectively. Curved texts are actually more prevalent in natural scenes and are more challenging for detection. Therefore, arbitrary shape text detection has attracted ever-increasing interest in research and industrial communities recently. Due to the robustness to shape variances, the Connected Component-based (CC-based) methods~\cite{SegLink++,CRAFT, DRRG} and segmentation-based methods~\cite{PSENet_v2, DB, TextField} gradually become the mainstream methods for arbitrary shape text detection. CC-based methods~\cite{SegLink++, TextDragon, CRAFT, DRRG} need complex processing for separating text instances into individual text patches during training and rely on complicated post-processing steps to aggregate candidate text patches into final text instances during testing. Due to the scalability of pixel-level prediction, segmentation-based methods can well adapt to texts with various shapes without complicated  processing and post-processing steps. So far, a series of promising segmentation-based arbitrary shape text detectors have been proposed, to name a few~\cite{CVPR19_LSA, CVPR19_PSENet, PSENet_v2, TextField, Boundary, DB}.

Different from generic object instance segmentation datasets, the existing text detection datasets only provide coarse-grained boundary annotations, as shown in Fig.~\ref{fig:fg1}. In annotations of text detection, there are a lot of background pixels, especially in the areas adjacent to boundaries, as shown in Fig.~\ref{fig:fg1}~(c) and Fig.~\ref{fig:fg2}. These ambiguous pixels may significantly degrade the performances of detection methods based on pixel-level prediction. Some segmentation-based techniques try to reduce or avoid the dependence on pixel prediction through distance field~\cite{MSR} or direction field~\cite{TextField}. However, the performances of these methods are still not satisfactory in curved text detection due to the unpredictable range of distances and directions. Although PSENet~\cite{CVPR19_PSENet} adopts a series of binary masks with different scales, its performance is still limited for the heavy dependence on the pixel classification results. In a text instance, the probability of whether a pixel belongs to text or not is highly related to the distance with its adjacent annotation boundary. The probability map can well describe text pixels' probability distribution, perfectly express the shape of text, and has a closed range of values. Therefore, it is more reasonable to use a probability map to describe a text than a binary mask or distance map. However, one probability map can not cover complex probability distributions well because of the uncertainty of coarse-grained text boundary annotations. 

Recently, some methods have adopted iterative modules to refine pixel predictions, such as the semantic segmentation method CANet~\cite{CANet} and the text detection method LOMO~\cite{CVPR19_LOMO}. In~\cite{CANet}, an iterative optimization module is adopted to refine predicted results for semantic segmentation. In LOMO~\cite{CVPR19_LOMO}, an iterative refinement module is adopted to refine the quadrangle proposals by regressing the coordinate offsets once or more times. These iterative models produce a prediction for the same target in every iteration step. However, in our method, we need to predict different probability maps in each iteration step and optimize the predictions of probability maps by implicitly learning the mapping relationship between successive probability distributions.

In this paper, we propose an innovative segmentation-based method via probability maps for accurately detecting text instances. Specifically, we adopt the Sigmoid Alpha Function (SAF) to transfer the distances between pixels and boundaries to different probability distribution maps by controlling a hyper-parameter in SAF. To further optimize the predictions of probability maps, we propose an iterative module (IM) equipped with asymmetric convolution kernels to optimize the predictions of probability maps by implicitly learning the mapping relationships between successive probability distributions. In this way, the high precise predictions of probability maps can be generated. Finally, some simple region growth algorithms (\eg, Watershed Algorithm or Progressive Scale Expansion Algorithm~\cite{CVPR19_PSENet}) can be used to aggregate probability maps to complete text instances. In addition, we design a voting-based filtering algorithm according to the properties of probability maps, which can reduce the hyper-parameters and improve the robustness of all datasets. Experiments on several challenging datasets demonstrate the excellent performance of our method.

In summary, our main contributions are three-fold:
\begin{itemize}
	\item We propose a novel and robust segmentation-based text detection method for arbitrary shape text detection, which employs probability maps to segment text instances accurately.
	
	\medskip
	\item We propose a flexible sigmoid alpha function (SAF) to establish the mapping between the pixel probability and the distance to boundary, which can generate different probability distribution maps by simply controlling a hyper-parameter.
	
	\medskip
	\item The proposed method obtains state-of-the-art performance in terms of detection accuracy both on polygonal and quadrilateral datasets. Specifically, our method with Watershed Algorithm achieves the best F-measure on Total-Text (88.79\%), CTW1500 (85.75\%), and MSRA-TD500 (88.93\%).
	
\end{itemize}

The rest of this paper is organized as follows. We briefly review related works on scene text detection in Section~\ref{Related_Work}.
The proposed method is elaborated in Section~\ref{Proposed_Method}, followed by extensive experiments in Section~\ref{Experiments}. Finally, we conclude and give some perspectives in Section~\ref{Conclusion}.

\begin{figure*}[htbp]
	\begin{center}
		\includegraphics[width=0.99\linewidth]{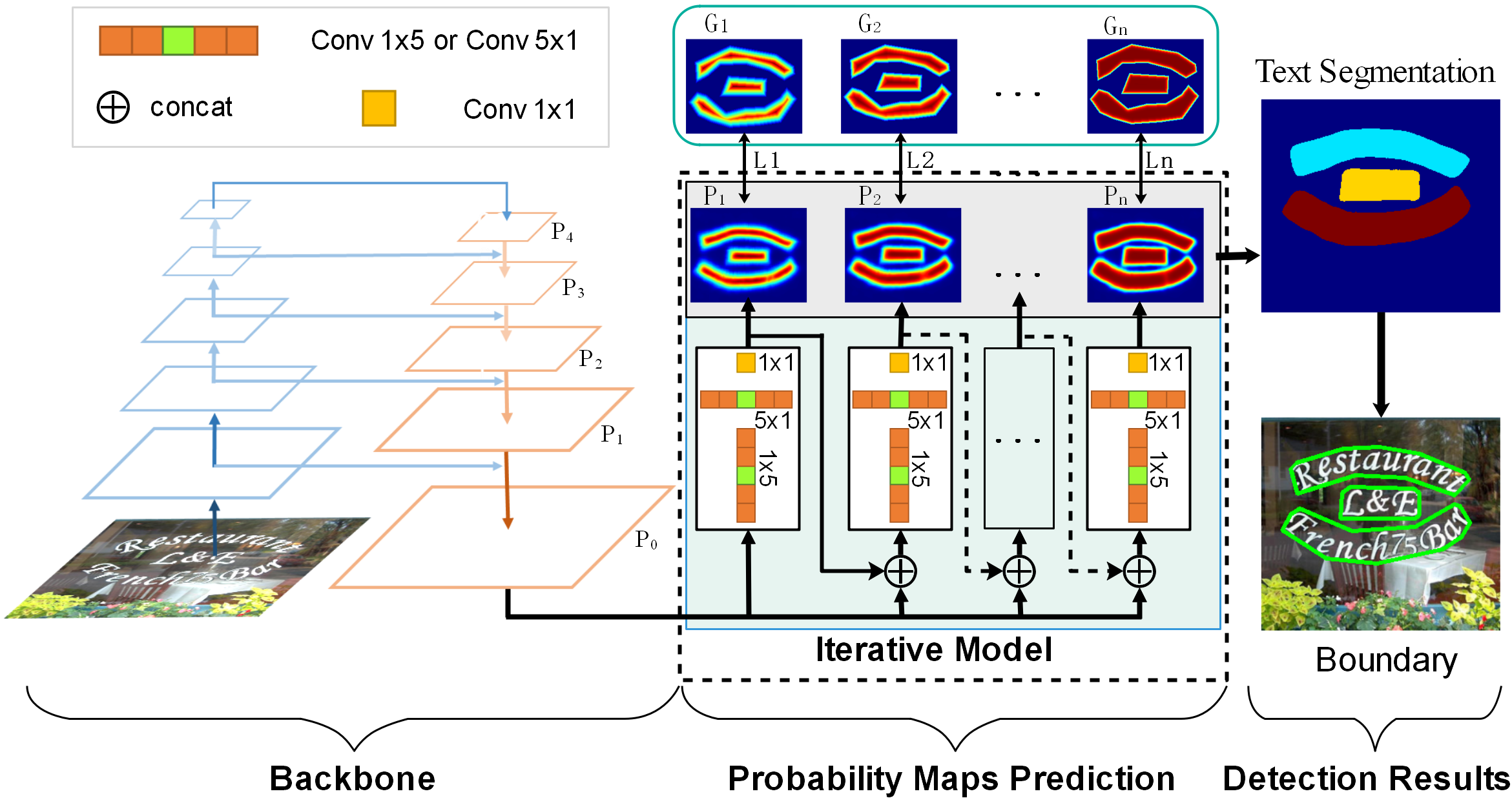}
		\caption{Framework of the proposed method. The backbone extracts the shared features, and 16 chancel fusion features are fed into the iterative model to iteratively generate the prediction of probability maps. }
		\label{fig:fg3}
	\end{center}
	\vspace{-1.2em}
\end{figure*}

\section{Related Works} \label{Related_Work}
Before deep learning flourished,
scene text detector pipelines usually consisted of text component extraction and filtering,
component grouping, and candidate filtering. The key step is extracting text components based on some engineered features. During this period, Maximally Stable Extremal Regions (MSER)~\cite{MSER04, MSER14, MSER15} and Stroke
Width Transform (SWT)~\cite{SWT10, SWT15} are two types of mainstream methods for text component extraction. With the prosperity of deep learning, convolutional neural network (CNN) based scene text detection methods have made great progress and substantially outperformed traditional MSER or SWT based text detection methods in terms of both accuracy and capability. In general, deep learning-based text detection methods can be subdivided into four categories: regression-based methods, CC-based methods, segmentation-based methods, and other methods.

\subsection{Regression-based Methods} This type of method relies on a box-regression based object detection frameworks with word-level and line-level prior knowledge~\cite{Zhang2022GraphFN, RRPN, DDR, textboxes++, EAST,Zhang2021AdaptiveBP}. Different from generic objects, texts
are often presented in irregular shapes with various aspect ratios. To deal with this problem, RRPN~\cite{RRPN} and Textboxes++~\cite{textboxes++} localize text boxes by predicting the offsets from anchors. Different from these methods localizing text regions by implementing refinement on pre-defined anchors, EAST~\cite{EAST} and DDR~\cite{DDR} propose a new approach for accurate and efficient text detection, which directly regress the offsets from boundaries or vertexes to the current point. Based on these direct regression methods, LOMO~\cite{CVPR19_LOMO} proposes
an iterative refinement module to iteratively refine bounding box proposals for extremely long texts and then predicts center line, text region, and border offsets to rebuild text instance. Although regression-based methods have achieved good performance in quadrilateral text detection, they often can't adapt well to arbitrary shape text detection.

\subsection{CC-based Methods}	The Connected Component (CC) based methods~\cite{CRAFT, DRRG, CTPN, SegLink} generally link or group the detected individual text parts or characters into final text instances by post-processing procedure. CTPN~\cite{CTPN}
modifies the Faster R-CNN~\cite{Faster-rcnn} to extract horizontal text components
with a fixed-size width for easily connecting dense text components and generating horizontal text lines. SegLink~\cite{SegLink} decomposes each text into two detectable elements, namely segment and link, where the link connects a pair of adjacent segments that belong to the same word. CRAFT~\cite{CRAFT} detects text regions by exploring affinities between characters. Zhang \etal~\cite{DRRG} 
use a graph convolution neural network (GCN) to learn and infer the linkage relationships of text components to group text components. CC-based methods have a more flexible representation and can adapt well to irregular shape text. Therefore, the CC-based methods are popular in arbitrary-shaped text detection, even if the clustering of components is unsatisfactory.

\begin{figure*}[htbp]
	\vspace{-0.6em}
	\begin{center}
		\includegraphics[width=0.98\linewidth]{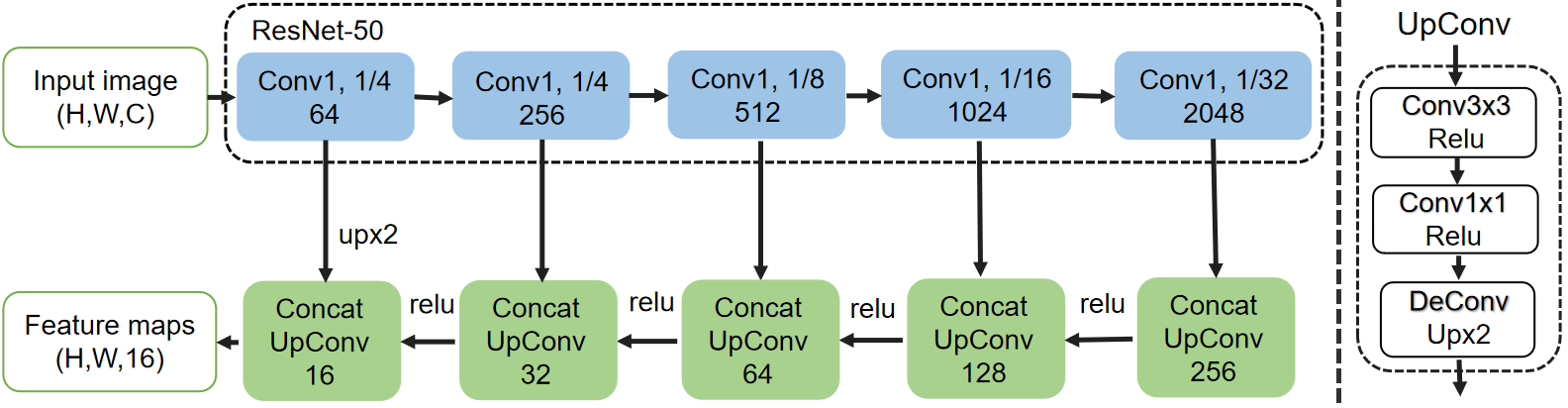}
		\caption{Architecture of the backbone. To capture multi-scale text instances, we adopt ResNet-50 as the backbone which is combined with a multi-level fusion strategy. The up-sample is performed by deconvolution. The 1/4, 1/8, 1/16, and 1/32 represent different scales of the feature maps. The 16, 32, 64, and 256 represent the number of output feature maps. The H, W, C represent the width, height, and the number of channels of input image, respectively.}
		\label{fig:fg4}
	\end{center}%
	\vspace{-1.2em}
\end{figure*}

\subsection{Segmentation-based Methods}
Methods~\cite{PixelLink,CVPR19_PSENet,CVPR19_LSA,TextField,TextSnake,Zhang2022KernelPN} in this type mainly draw inspiration from semantic segmentation methods and detect texts by estimating word bounding areas. Segmentation-based methods generally use different representations to describe text regions and then rebuild text instances through specific post-procession.
In PixelLink~\cite{PixelLink}, linkage relationships between a pixel and its neighboring pixels are predicted for grouping pixels belonging to the same instance. In~\cite{CVPR19_LSA}, embedding features are used to provide instance information and to generate a prediction for text instances of arbitrary shape.
To effectively distinguish adjacent text instances, PSENet~\cite{CVPR19_PSENet} adopts a progressive scale expansion algorithm to expand the pre-defined kernels gradually.  However, the performances of these methods are strongly affected by the quality of segmentation accuracy.

\subsection{Other Methods}
In addition to the above methods, here are some other methods. The end-to-end methods~\cite{MaskTextSpotter, TextDragon, FOTS, Boundary} unify text detection and recognition into an end-to-end network. In general, these end-to-end approaches can achieve higher detection performance with the help of text recognition information, such as~\cite{Boundary}.
Besides, Yao \etal~\cite{corner} detect texts by predicting the corner of the texts, and  Lyu et al.~\cite{CVPR19_ATRR} adopt a similar architecture as SSD~\cite{SSD} and rebuild text instances with predicted corner points.

\begin{figure}[htbp]
	\centering
	\subfigure[]{
		\begin{minipage}[t]{0.48\linewidth}
			\centering
			\includegraphics[width=4.4cm,height=2.7cm]{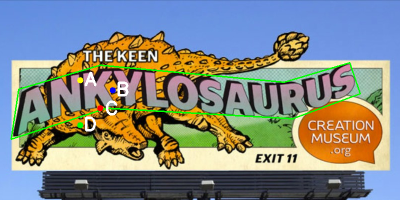}
		\end{minipage}
	}%
	\subfigure[]{
		\begin{minipage}[t]{0.48\linewidth}
			\centering
			\includegraphics[width=4.4cm,height=2.8cm]{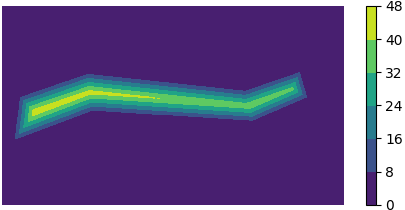}
		\end{minipage}
	}%
	\vspace{-4pt}
	\quad
	\subfigure[]{
		\begin{minipage}[t]{0.98\linewidth}
			\centering
			\includegraphics[width=8.7cm,height=4cm]{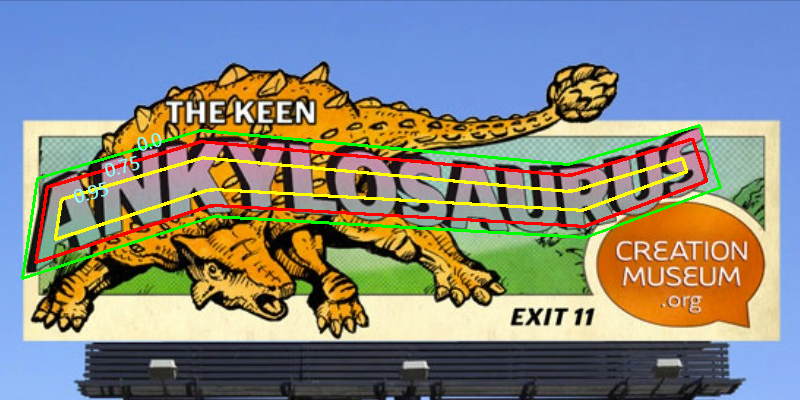}
		\end{minipage}%
	}%
	\caption{Illustration of the relation between probability distribution and the distance to the boundary. (a) Some sampling points and boundaries (green contour) are annotated in labels, (b) Distance map, and (c) Boundaries of different sizes, where the green contour is annotated in labels. The number on the contour represents its probability value.}\label{fig:fg5}
	\centering
	\vspace{-1.2em}
\end{figure}

\section{The Proposed Method}\label{Proposed_Method}
\subsection{Overview}
Fig.~\ref{fig:fg3} illustrates the framework of our method. We adopt ResNet-50 ~\cite{ResNet} to extract features. To preserve spatial resolution and take full advantage of
multi-level information, we exploit a widely used multi-level feature fusion strategy (similar to U-Net~\cite{UNet}). Fig.~\ref{fig:fg4} illustrates the architecture of our backbone. An iterative module (IM) is used to produce the predictions of corresponding probability maps based on feature maps extracted by the backbone to optimize the results. The IM also refines predictions of probability maps by implicitly learning the mapping relationships between successive probability distributions. These predicted probability maps can describe the possible probability distributions of text pixels within a boundary and provide enough information for reconstructing text instances. 
We adopt different region growing algorithms to aggregate the computed probability distribution maps into candidate text instances. Finally, we adopt an adaptive voting filtering algorithm to filter out false-positive text instances.

\subsection{Probability Maps}
\label{Probability_Maps} 
Because of the extremely high cost of pixel-level annotating, scene text detection datasets only provide coarse-grained boundary annotations. Most segmentation-based methods~\cite{PSENet_v2, CVPR19_LSA, CVPR19_PSENet, DB} directly convert these coarse-grained annotations into pixel-level binary supervision masks for text instance segmentation. However, the quality of generated pixel-level labels is always unsatisfactory. As shown in Fig.~\ref{fig:fg1}~(c), there exist a lot of background pixels, mainly those pixels adjacent to boundaries. Many background pixels not belonging to text regions may be incorrectly categorized into text pixels in these pseudo-binary masks. If trained on noisy data, segmentation-based text detectors tend to get false-positive text instances. Although some segmentation-based methods try to reduce or avoid the dependence on binary masks through distance field~\cite{MSR} or direction field~\cite{TextField}, the inaccurate annotations around text boundaries may significantly degrade the detection performance.

In our work, we propose probability maps to solve the uncertainty in text annotations explicitly. Generally speaking, the probability of whether a pixel is a text pixel is related to the distance to the boundary.  As shown in Fig.~\ref{fig:fg5}~(a), the green contour is an annotated boundary, the red point ($ C $) is near the boundary, and the yellow point ($  A $) is away from the text boundary. For text detectors, the probability of predicting $ A $ as a text pixel is much greater than that of $ C $, because $ A $ is farther away from the boundary than $  C $. Although the blue point ($ B $) is more like a background pixel only considering color and texture features, the probability that $ B  $ is predicted as a part of a text instance is still very high in the existing text detection methods. The CNN-based text detection methods, \eg \cite{PixelLink, CVPR19_CSE, DB}, use not only simple features such as color and texture but also use the information of other pixels in the receptive field. Point $ B $ is far away from the text boundary and surrounded by text pixels, so the probability of $ B $ predicted as a text pixel is very high in text detectors.

As our ultimate goal is to get the text boundaries, it is more reasonable to consider the probability of the whole boundary instead of a single pixel. So in our work, we think that pixels on the same contour have the same probability, and the contours closer to the annotation boundary have the smaller probability value. As shown in Fig.~\ref{fig:fg5}~(c), the green contour is an annotation boundary, which is basically in the background region. And there are also a lot of background noises inside the boundary, especially near the boundary, such as point $ C $ in Fig.~\ref{fig:fg5}~(a). The pixels on the red contour have the same distance to the text boundary (green contour), and the pixels on the yellow contour also have the same distance to the text boundary. Compared with the green contour, the pixels on the red contour are more likely to be a text instance because the proportion of text pixels on the red contour is higher than that of the green contour. Similarly, pixels on the yellow contour are more likely to be text than pixels on the red and green contours. By calculating the distance to annotation boundary, we can get a series of contours on which pixels have the same distance to annotation boundary, as shown in Fig.~\ref{fig:fg5}~(b). In general, the contour is closer to the boundary, the proportion of text pixels is lower. Suppose we use the proportion of text pixels on the contour as pixel probability. In that case, the pixels on the same contour will have the same probabilities, which is convenient for us to establish the relationship between the probability of a pixel and its distance to the boundary. For the text detection task, our goal turns to explore a contour on which the probability of pixels has a minimum value (approximating to zero).

\begin{figure}[ht]
	\begin{center}
		\includegraphics[width=0.98\linewidth]{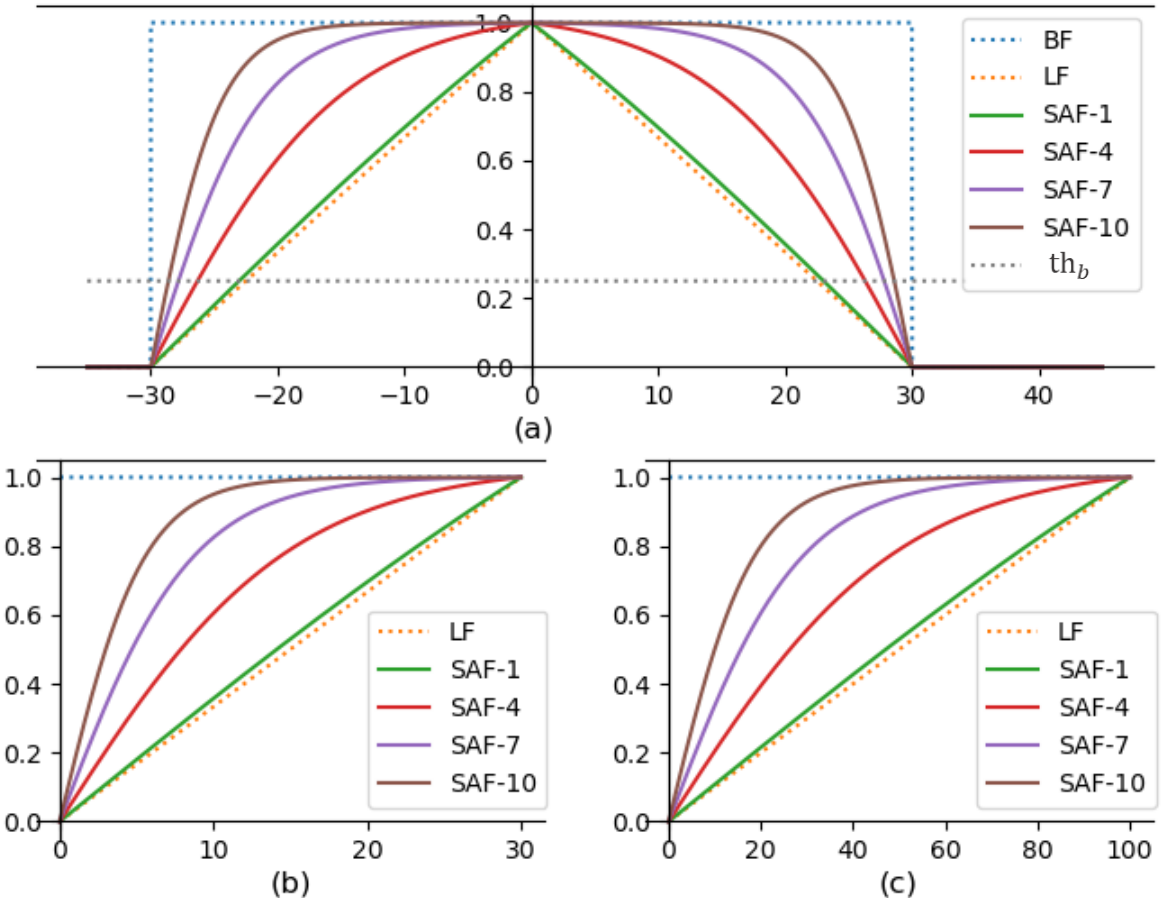}
		\caption{Illustration of different probability distribution curves generated by $ SAF $, $ BF $ and $ LF $: (a) Cross-section curves of different mapping function for text instance, (b) Probability distribution curves for text instance with scale 30, and (c) Probability distribution curves for text instance with scale 100. The  $ th_b $ refers to binary threshold in Algorithm~\ref{algo:threshold} and Algorithm~\ref{algo:vote}.The horizontal axis represents the distance from the text pixel to the boundary, and the vertical axis represents the probability value.
		}
		\label{fig:fg6}
	\end{center}%
	\vspace{-1.5em}
\end{figure}

In this work, we build a probability map based on the distance between pixels and annotation boundary to describe the probability distributions where pixels inside a boundary belong to a text. To establish the mapping between the pixel probability and the distance to boundary, we propose a  Sigmoid Alpha Function ($ SAF $) based on the Sigmoid Function \cite{Sigmoid}. Because the Sigmoid Alpha Function can map a distance value into range $ [0,1] $, which can be regarded as a pixel probability. Hence we can adopt $  SAF $ to map a distance map ($ D $) into a probability map. The Sigmoid Alpha Function ($ SAF $) is defined as:
\begin{gather}
SAF_{(i,j)} = C*(\dfrac{2}{1+e^{\frac{-\alpha D_{(i,j)}}{L}}} - 1), \label{SAF}\\
C = \frac{1+e^{-\alpha}}{1-e^{-\alpha}}; \;\; \alpha \in (0, \infty),
\end{gather}
where $ \alpha  $ is a transforming factor for generating different probability distribution maps; $ C $ is a constant for keeping the value range of the Sigmoid Alpha Function in $ [0,1] $ and  ensuring the probability of the farthest pixel to boundary is 1; ($ i,j $)  indicates the coordinate point in the map; $ D_{(i,j)}  $ is the shortest distance from pixel $ (i,j) $ to boundary; distance map ($ D $) is the set of  $ D_{(i,j)}  $; $  L $ represents the scale of text instance, and is defined as:
\begin{equation}
L = max(D_{(i,j)}); \quad i,j \in T,
\end{equation}
where $  T $ represents a text instance. $ L $ in $ SAF $ encodes the size information of text instances into probability maps, improving the model adaptability in detecting text with varied sizes.

The visual illustrations of different probability distribution curves generated by $ SAF $ are shown in Fig.~\ref{fig:fg6}. The  proposed $ SAF $ has two special cases, $ LF $ and $ BF $:
\begin{gather}
LF_{(i,j)} = D_{(i,j)}/L;  \; D_{(i,j)} \ge 0, \label{LN}\\
BF_{(i,j)} =
\begin{cases} 
1,\; if \;  D_{(i,j)} \textgreater th,\\
0, \; otherwise,
\end{cases} \label{BN}
\end{gather}
where $ LF $ is a linear normalization function, $ BF $ is a binarization function and  $ th $ is a pre-defined threshold. As shown in Fig.~\ref{fig:fg6}, if $ \alpha = 1 $, the curve of $ SAF $ is very close to the curve of $ LF $. Along with $ \alpha $ increases, the curve of $ SAF $ tends to the curve of $ BF $. The mathematical formulations are elaborated as:
\begin{gather}
LF_{(i,j)} = {\lim_{ \alpha \to 0}} \, SAF_{(i,j)}(\alpha), \label{eq1}\\
BF_{0_{(i,j)}} = {\lim_{ \alpha \to +\infty}}SAF_{(i,j)}(\alpha),\label{eq2}
\end{gather}
where $ BF_{0_{(i,j)}} $ is a special case of Eq.~\ref{BN} if its threshold ($ th $) is set to zero. From Eq.~\ref{eq1} and Eq.~\ref{eq2}, we can conclude that both the linear normalization function $  LF $ and the binarization function $ BF $ are two special cases of $ SAF $. Therefore, the binary mask and the linear normalized distance map can be regarded as two special probability distribution maps. The binary mask is widely used in segmentation-based text detection methods, in which all pixels within the annotation boundary are regarded as text pixels, ignoring background noises. It is usually difficult to achieve superior performance only with binary masks as training supervision~\cite{CVPR19_PSENet, PixelLink}. Some other  methods~\cite{DB, PSENet_v2, TextSnake} adopt additional information to supervise network learning. The normalized distance map with $  LF $ is usually used as auxiliary information for text instance segmentation~\cite{PMTD, TextMountain, DB}. And the normalized distance map with $  LF $ can be regarded as a probability map of the linear distribution of text-pixel probabilities inside the boundary.

Because of the uncertain and inaccurate annotation of boundaries, it is impossible to accurately describe the probability distributions of text pixels inside a boundary only with one probability map. To segment text instances accurately, the works in ~\cite{DB, TextField, TextSnake} adopt a variety of prediction information to detect text instances. In our work, we employ a series of probability distribution maps as supervision to segment text instances. According to our observation and experiments, a group of probability maps can describe possible probability distributions for covering annotation error and providing abundant information for reconstructing text instances. 
In addition, with multiple probability maps, our model can be more robust for final detections. Specifically, even if a local prediction of one probability map is not accurate, there are still several other probability maps that can provide supplementary context predictions. According to Eq.~\ref{SAF}, we can easily get a group of probability distribution maps by controlling the hyper-parameter $ \alpha $ in $ SAF $, including linear and binary probability distributions.

\textbf{Label Generation:} The label generation procedure of the probability maps relies on the distance map ($ D $) and $ SAF $. Given a text image, each text instance region is described by a polygon boundary. Different datasets are usually labeled with polygons with different vertex numbers, four for quadrilateral text datasets and 14 for CTW1500 or any vertexes for Total-text. For each text instance, we first compute the shortest distance $  D_{(i,j)} $ from pixel $ (i,j) $ to boundary. So, the distance map of each text instance is described by a set of $  D_{(i,j)} $ as
\begin{equation}
D = \{ D_{(i,j)} \};\quad i,j \in T	
\end{equation}

\begin{figure}[htbp]
	\vspace{-0.6em}
	\begin{center}
		\includegraphics[width=0.9\linewidth]{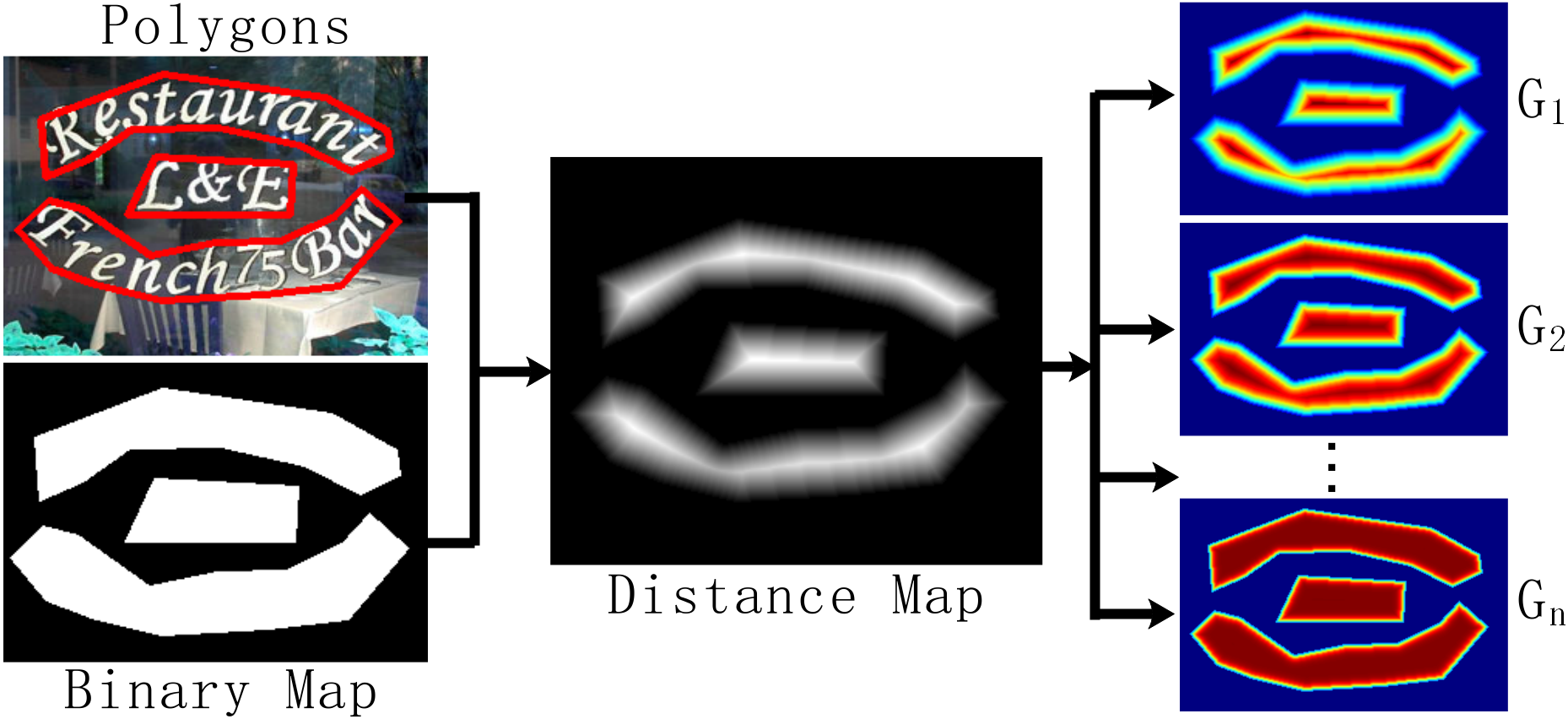}
		\caption{ The generation of different probability map labels , where the annotation of text polygon is visualized in red lines, and $ \{G_0, G_1,...,G_n\} $ are the corresponding results using $ SAF $ with different alphas.}
		\label{fig:fg7}
	\end{center}
	\vspace{-1.2em}
\end{figure}

After getting the distance map ($ D $), we use the sigmoid alpha function with a set of $ \alpha $ values ($ \alpha_1,\alpha_2,...,\alpha_n $) to map the distance into probability, as shown in Fig.~\ref{fig:fg7}. The probability maps can be described by a set of probability distribution, \ie, 
\begin{gather}
G =\{ G_i \}_{k=1}^{n}=\{ SAF(D, \alpha_i)| i \in (1,2,...,n) \}\\
p_{i(p)} =\{ SAF(D_{(p)}, \alpha_i)|i \in (1,2,...,n) \}
\end{gather}
where $ n $ is the number of $ \alpha $ values and probability maps. These generated probability maps ($ G $) will be used as the final ground-truths. When generating probability maps ($ G $), the maximum value in the possible overlapping region of probability maps will be selected. The $ p_{i(p)} $ is the probability of pixel $ p $ with $ \alpha_i $ in $ SAF $.

\textbf{Alpha ($ \alpha $) Selection:} For facilitating our experiments, the set of $\{ \alpha_{1},\alpha_{2},...,\alpha_{n} \} $ are only selected from positive integers.  The first term $ \alpha_{1} $ is equal to 1 ($ \alpha_{1}=1 $), and the step is $ k $ ($  \alpha_{i} = \alpha_{i-1}+k$).
In details, the value of $ \alpha_{i} $ in $ i $-th iteration can be calculated by
\begin{gather}
\alpha_{i} = k*(i-1)+1; i\in [1,n], k \in [2,\infty),\label{alpha_select}
\end{gather}
where $ n $ is the number of probability maps; $ k $ is the step of $ \alpha $ value; $ i $ indicates the $ i $-th probability map. For example, when $ k=2, n=4 $, four probability maps will be generated, and four corresponding alpha values are $\{\alpha_{1}=1,\alpha_{2}=3,\alpha_{3}=5, \alpha_{4}=7 \}$. In our experiments, the value of $ \alpha $ always starts from 1, because the probability distribution is close to the linear distribution when $ \alpha = 1$, facilitating distinguishing different text instances. 
The number of probability maps is at least two. The interval of $ \alpha $ is at least two because one probability map can not provide enough information for reconstructing text instances.

\begin{figure}[htbp]
	\begin{center}
		\includegraphics[width=0.95\linewidth]{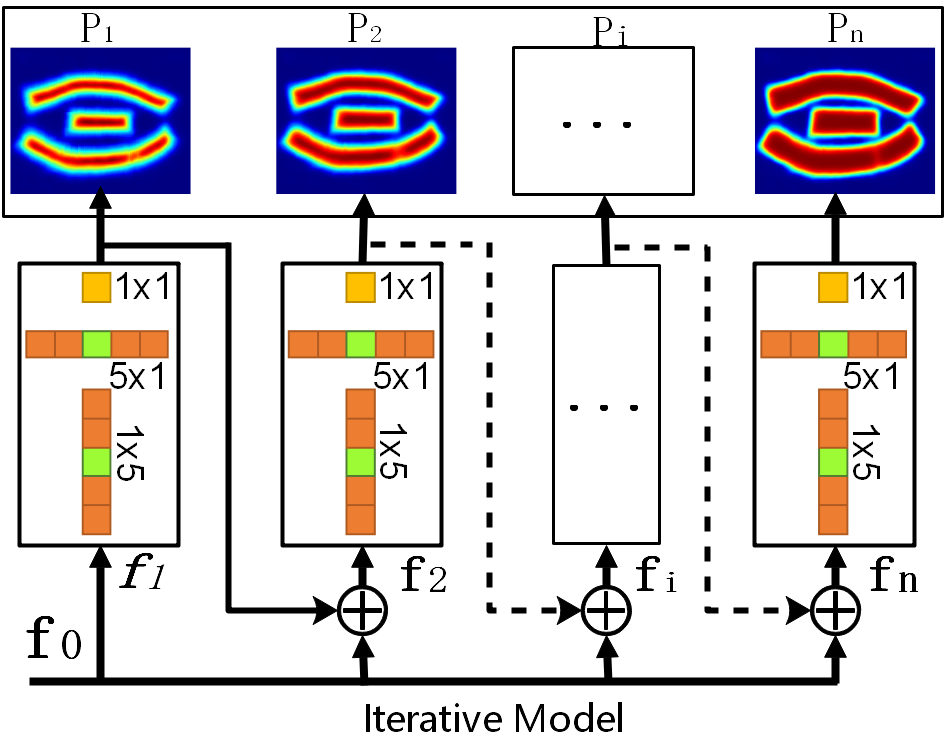}
		\caption{The structure of iterative module, where $ f_0 $ represents the feature extracted by backbone, and the set $ \{P_1,P_2,...,P_i,...,P_n\} $is the prediction of probability maps.}
		\label{fig:fg8}
	\end{center}
	\vspace{-1.2em}
\end{figure}

\subsection{Iterative Module}
\label{IM}

The iterative structure is prevalent in segmentation methods, which can significantly improve detection accuracy. In CANet~\cite{CANet}, Zhang \etal adopted an iterative module to optimize the predicted results for semantic segmentation. In LOMO~\cite{CVPR19_LOMO}, an iterative refinement module is also used to refine quadrangle proposals by regressing the coordinate offsets. 

In this work, we propose an innovative iterative module (IM) with large-scale asymmetric convolution kernels determined by text attributes. The detailed structure of our IM is shown in Fig.~\ref{fig:fg8}. Different from the existing iterative modules ~\cite{CANet,CVPR19_LOMO}, the predictions of our IM are targeted for a series of ground-truth probability maps. In this way, our IM can further optimize the predictions of probability maps by exploring the mapping relationship among probability distributions. With the help of our IM, our network can fully utilize the information of predictions from low-level iterations to improve the accuracy of predictions at high-level iterations. And the parameters in high-level iteration layers can be optimized by a back-propagation gradient from the layers at high-level iterations. 

Different from generic object instances, scene text instances have significantly varied sizes and aspect ratios and are usually distributed in strips, especially for long text instances and line-level text instances. To solve this problem, Pixel-Anchor~\cite{Pixel-Anchor}  and IncepText~\cite{IncepText} adopt the large-scale asymmetric convolution kernels, and experiments show that this strategy is effective. 
In our work, we solve the problem of significantly varied sizes and aspect ratios in scene text detection from two aspects. On the one aspect, we use $ SAF $ to encode text size information 
into probability maps. On the other aspect, inspired by ~\cite{Pixel-Anchor} and~\cite{IncepText}, we use a group of asymmetric convolutions in our iterative model to improve the robustness for long text and line-level text. Equipped with the above-mentioned two strategies, our method can be commendably adapted to detect long text instances. Mathematically, ou IM can be formulated as:
\begin{equation}
f_1 = ConvBlock(f_0); \; P_1 = \sigma(f_1),
\end{equation}
\begin{equation}
f_i = ConvBlock(f_0 \oplus f_{i-1}),
\end{equation}
\begin{equation}
P_i = \sigma(f_i); \quad i\in (1, n],
\end{equation}
where ``$ \oplus $'' refers to the concatenation operation; $ \sigma(\cdot) $
refer to the $ Sigmoid $ activation function; $ ConvBlock $ consists of
a group of asymmetric convolutions (\eg, $ Conv_{1 \times 5} $, $ Conv_{5 \times 1} $ and $ Conv_{1 \times 1} $); $ n $ denotes the number of iteration steps.

After getting fusion feature maps ($ f_0 $) of 16 channels from the backbone, we feed $ f_0 $ into the first iterative layer of our IM to produce feature map $ f_1 $. Then, $ f_1 $ will be used to generate a probability map through the $ Sigmoid $ function and will be fed into the next iteration layer to provide prior information for producing the second probability map. In this way, we can get a group of probability map predictions. From our experiments, we observed that the accuracy of text instance segmentation will be gradually improved with the increase of iterations. 


\subsection{Optimization}

\subsubsection{Loss Function}
\label{subsubsec:trainingobjective}
As shown in Fig.~\ref{fig:fg4}, we leverage backbone to extract shared features, and then adopt our IM proposed in Section.~\ref{IM} to generate the predictions of corresponding probability maps.  Each probability map has its own independent individual loss computed by Mean Squared Error (MSE) of each pixel on image domain $ \Omega $, and computed as:
\begin{gather}
loss_i =  \dfrac{1}{\Omega}\sum_{p \in \Omega}||G_i(p),P_i(p)||_2
\end{gather}
where $ G_i $ is the $ i $-th ground-truth of the probability map, and $ P_i $ is the $ i $-th predicted probability map; $ p $ denotes pixel in the image domain $ \Omega $.

The total loss $ L_{s} $ in our method is the weighted sum of individual losses ($ loss_i $), \ie, 
\begin{gather}
L_{s}= \sum_{i=1}^{n}\lambda_i*loss_{i},
\end{gather}
where $ loss_{i} $ is the loss of the $ i $-th probability map; $ n $ is the number of probability maps; $ \lambda_i $ is the weight of each individual loss, and all weight ($ \lambda $) are set to 1 in our experiments.

\vskip 1mm
\subsubsection{Online Hard Negative Mining}
\label{subsubsec:OHEM}
In scene images, text instances usually occupy a small area. Thus, the
number of text pixels and non-text pixels is rather imbalanced. To make the network training
focus more on pixels that are hard to distinguish, we adopt the online hard negative mining (OHEM) strategy proposed in~\cite{OHEM} for all regression losses $ \{ loss_1, loss_2,..., loss_n \}$. More specifically, non-text pixels with zero probability are sorted in decreasing order by their per-pixel losses. Then, only the high-ranked non-text pixels are reserved for $\gamma* (\sum_{p \in \Omega} \{p| p_{i(p)}>0\})$
back-propagation, where $\gamma$ is a given hyper-parameter that denotes the ratio of non-text pixels with respect to the total number of text pixels. In our experiments, $\gamma$ is empirically set to 3.

\subsection{Inference and Post-processing}
After we get the probability distribution maps $ \{P_1, P_2, ..., P_n\} $ from the trained network, some simple region growth algorithms (\eg, Watershed Algorithm or Progressive Scale Expansion Algorithm~\cite{CVPR19_PSENet}) will be used to aggregate these probability maps to complete text instance. As a prerequisite, we should binarize all probability maps with the same small constant binary threshold ($ th_{b} $). By reconstructing for text region, we will get a series of candidate text instances $ \{S_1, S_2, ..., S_m\} $, and each text instance ($ S_i $ ) is represented by a connected binary mask.

\begin{algorithm}
	\caption{Threshold Filtering Algorithm}
	\label{algo:threshold}
	\LinesNumbered
	\SetAlgoLined
	\SetKw{kAnd}{and}
	\SetKw{kOr}{or}
	{\em \textbf{Require:} }\\
	\hspace{2em}Average Probability Threshold: $ th_e $,\\
	\hspace{2em}Text Region Min Area:  $\beta_a $,\\
	\hspace{2em}Predicted Probability  Map: $ P_n $,\\
	\hspace{2em}Text  Region Binary Masks: $ \{S_1, S_2,...,S_m\} $;\\
	$ \mathcal{E} \gets 0 $, $ \mathcal{N} \gets 0 $, $\mathcal{R} \gets \emptyset$,   //initialization \;
	//{\it Filter small region text}\\
	//{\it Filter low average probability text}\\
	\ForEach{$S_i \in \{S_1, S_2,...,S_m\}$}
	{
		$ \mathcal{E} \gets reduce\_mean(P_n, mask \gets S_i) $;\\
		$ \mathcal{N} \gets reduce\_sum(S_i)$;\\
		\If{$\mathcal{E} \geq th_e$ \kAnd $ \mathcal{N} \geq \beta_a$ } 
		{
			$\mathcal{R} \gets \mathcal{R} \cup S_i$;
		}
	}
	\Return{$\mathcal{R}$} \;	
\end{algorithm}

\textbf{Text instance filtering:} After extracting candidate text instances $ \{S_1, S_2, ..., S_m\} $, we apply two filtering strategies (threshold filtering and voting filtering) to get rid of non-text instances according to probability and size. Generally speaking, the predicted value of the text probability map should be close to the actual value, and the predicted value of the non-text probability map is usually tiny overall. So, we can filter out non-text instances according to their probability mean values to improve detection performance. 	Besides, we also discard some noisy candidate instances whose areas are smaller than $\beta_a $. Finally, the remaining candidate text instances are the final predictions.

\vskip 1mm
\subsubsection{Threshold Filtering}
Although the generic threshold filtering algorithm is simple and fast, it is sensitive to the pre-defined threshold and hence not robust across different datasets. Therefore, it is necessary to adjust the threshold appropriately for different datasets. Specifically, our threshold filtering algorithm is detailed in Algorithm~\ref{algo:threshold}. 
We calculate an average probability and an area for each candidate text instance. Then, we eliminate the candidate text instance with low average probability and too small area (smaller than 300 pixels) through the pre-defined threshold. There are two thresholds ($ th_b $ and $ th_e $) that need to be empirically adjusted on different datasets for the threshold filtering algorithm. 

\begin{algorithm}
	\caption{Voting Filtering Algorithm}
	\label{algo:vote}
	\LinesNumbered
	\SetAlgoLined
	\SetKw{kAnd}{and}
	\SetKw{kOr}{or}
	{\em \textbf{Require:} }\\
	\hspace{2em}Binary Threshold: $ th_{b} $,\\
	\hspace{2em}Text Region Min Area:  $\beta_a $,\\
	\hspace{2em}Predicted Probability  Maps: $ \{P_1, P_2, ..., P_n\} $,\\
	\hspace{2em}Text  Region Binary Masks: $ \{S_1, S_2,...,S_m\} $,\\
	\hspace{2em}Voting weight: $ \{w_1, w_2,...,w_n\} $,\\
	\hspace{2em}Alpha Set: $ \{\alpha_1, \alpha_2,...,\alpha_n\} $;\\
	
	//initialization \;
	$ \mathcal{V} \gets 0 $, $ \mathcal{E}_{old}, \mathcal{E}_{new} \gets 0 $, $ \mathcal{N} \gets 0 $, $\mathcal{R} \gets \emptyset$;   \\
	//{\it Filter small  text  and low average probability text}\\
	\ForEach{$S_i \in \{S_1, S_2,...,S_m\}$}
	{
		$ \mathcal{N} \gets reduce\_sum(S_i)$;\\
		\lIf{$ \mathcal{N} \textless \beta_a $}
		{
			continue
		}
		\ForEach{$P_j \in \{P_1, P_2,...,P_n\}$}
		{
			
			$ \mathcal{E}_{old} \gets reduce\_mean(P_j, mask \gets S_i) $;\\
			$ D_i \gets compute\_distance\_map(S_i) $;\\
			$ P' \gets compute\_probability\_map(D_i, \alpha_j) $;\\
			$ \mathcal{E}_{new} \gets reduce\_mean(P', mask \gets S_i) $;\\
			\eIf{$ \mathcal{E}_{old} \geq \mathcal{E}_{new} - th_{b}^{2}$}
			{
				$ \mathcal{V} \gets \mathcal{V} + w_j*1 $;
			}
			{
				$ \mathcal{V} \gets \mathcal{V} + w_j*0 $;
			}
		}
		\If{$ \mathcal{V} \geq 0.5$}
		{
			$\mathcal{R} \gets \mathcal{R} \cup S_i$;
		}	
	}
	\Return{$\mathcal{R}$} \;	
\end{algorithm}

\vskip 1mm
\subsubsection{Voting Filtering}
The threshold filtering algorithm is sensitive to threshold parameters in testing because it uses fixed thresholds. Besides, it is not convincing to use only the last probability map as a decision basis. Hence, we propose a voting filtering algorithm, which can effectively improve the adaptability and robustness of our model to different datasets in testing. Specifically, the voting filtering algorithm is detailed in Algorithm~\ref{algo:vote}. After extracting binary masks $ \{S_1, S_2, ..., S_m\} $ of candidate
text instances, we firstly filter out the text with a small area (smaller than 300 pixels). Then we calculate the expectation of  each probability map ($ \mathcal{E}_{old} $) in the region corresponding to $ S_i $. At the same time, we use $ SAF $ to map $ S_i $ into a group of probability maps
with the same alpha ($ \alpha $) set as in training and calculate the expectation ($ \mathcal{E}_{new} $) of each newly generated probability map again. Theoretically, $ \mathcal{E}_{old} $ should be greater than $ \mathcal{E}_{new} $, but there are errors in the probability map of network prediction. Therefore, we set an error offset related to the threshold ($ th_b $). As shown in line 20 in Algorithm~\ref{algo:vote}, if this condition is met, the corresponding vote will be 1; otherwise, the vote will be 0. Each probability map will generate a unique voting result (0 or 1) with a voting weight. If the final weighted voting result is less than 50\%, the corresponding text instance $ S_i $ will be filtered out.

\textbf{Extract text boundaries:} Finally, we need to get text boundaries from the results of text instance segmentation. For curved text detection datasets, we use the alpha shape algorithm in OpenCV to extract the polygonal boundary of text instances. In contrast, we utilize the minimum enclosing rectangle of a text instance as its boundary for quadrilateral text detection datasets.

\section{Experiments}\label{Experiments}
We evaluate the proposed method on five public benchmark datasets: SCUT-CTW1500~\cite{CTW1500} and Total-Text~\cite{TotalText} which contain curved texts; ICDAR2015 Incidental Scene Text (ICDAR2015)~\cite{IC15}, ICDAR2017-MLT~\cite{MLT} and MSRA-TD500~\cite{MSRATD500} which mainly consist of multi-oriented texts. As a common technique, SynthText in the Wild~\cite{SynText} is also adopted to pre-train the network. Some implementation details are depicted
in Section~\ref{Implement-Details}, followed by ablation study experiments in Section~\ref{Ablation-Study}. The experimental results compared with state-of-the-art methods on  public datasets are given in Section~\ref{main_result}.

\subsection{Datasets}
\noindent\textbf{SynthText}~\cite{SynText}: It contains 800k synthetic images generated by blending natural images with artificial text. Annotations are given in types of characters, words, and lines. This dataset with word-level annotation is used to pre-train the proposed model.

\noindent\textbf{Total-Text}~\cite{TotalText}: It contains 1,555 scene images, divided into 1,255 training images and 300 testing images. This dataset is originally tailored for the arbitrary-shaped text detection, containing many curved and multi-oriented texts. Annotations are given in word level with polygon-shaped bounding boxes instead of conventional rectangular bounding boxes.

\noindent\textbf{CTW-1500}~\cite{CTW1500}: It consists of 1,000 training images and 500 testing images. Different from classical multi-oriented text datasets, this dataset is quite challenging due to many curved texts.  Each text instance is labeled by a polygon with 14 points. The annotation is given in line or curve level.

\noindent\textbf{MSRA-TD500}~\cite{MSRATD500}: It consists of 500 training images and 200 testing images, with English and Chinese scripts. This dataset is dedicated to detecting multi-lingual long texts of arbitrary orientations, annotated at the level of text lines. 

\noindent\textbf{ICDAR2015}~\cite{IC15}: It consists of 1,000 training images and 500 testing images. This dataset was released for the Challenge 4 of ICDAR2015 Robust Reading Competition, which is widely used as a benchmark dataset for multi-oriented text detection. Different from previous datasets with texts captured in relatively high resolution, scene images in this dataset are randomly acquired by Google Glasses. Therefore, texts in these images are of various scales, orientations, contrast, blurring, and viewpoints, making it challenging for detection.  Annotations are provided with word-level bounding quadrilaterals.

\noindent\textbf{ICDAR2017-MLT}~\cite{MLT}: It consists of 7,200
training images, 1,800 validation images, and 9,000 test images in nine languages. Similar to ICDAR2015, the text regions in ICDAR2017-MLT are also annotated by the four vertices of quadrilaterals. We use both the training set and the validation set in training.

\subsection{Implementation Details}\label{Implement-Details}
In our experiments, we first pre-train our network on the SynthText dataset by two epochs, in which images are randomly cropped and resized to $ 512 \times 512 $. In pre-training, the Adam~\cite{ADAM} optimizer is applied to train our model with a fixed learning rate of $ 0.001 $, and the batch size is 10. In fine-tuning, the training process is divided into two stages: In the first stage,  we randomly crop the text region, resize them to $ 640 \times 640 $ and train the model with batch size 6; In the second stage, we randomly crop the text region, resize them to $ 800 \times 800 $ and train our model with batch size 4. For both two stages, the SGD~\cite{SGD} optimizer is adopted in which the initial learning rate is $ 0.01 $ and multiplied by $ 0.9 $ after every 100 epochs. The data augmentation for the training data includes random rotation with an angle (sampled by Gaussian in $ (-60^{\circ}, 60^{\circ}) $), random cropping, and random flipping. In inference, we keep the aspect ratio of test images and resize them to the appropriate size for testing. Experiments are all performed on a single GPU (GeForce RTX-2080), Intel Xeon Silver 4108 CPU @ 1.80GHz, and PyTorch 1.2.0. 

\begin{figure}[htbp]
	\begin{center}
		\includegraphics[width=0.98\linewidth]{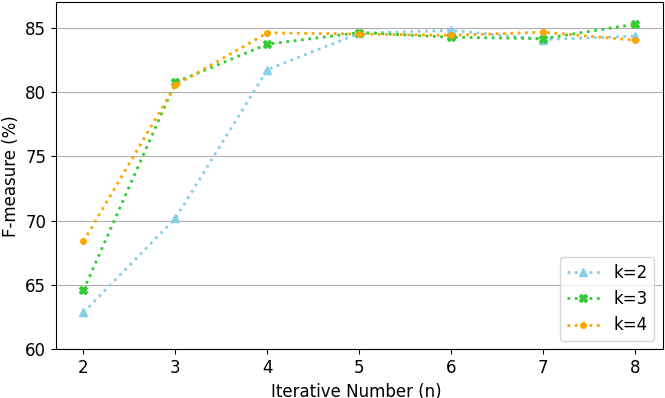}
		\caption{Ablation study on iterative steps (n) and the step of $ \alpha $ value (k). The results are only based on Total-Text dataset without pre-training.}
		\label{fig:carve_line}
	\end{center}
	\vspace{-1em}
\end{figure}

(Note: the ``our-1s” means that the output feature map is 1/1 of the input image, and the feature map $ P_0 $ shown in Fig.~\ref{fig:fg3} as the input of the iterative module. The ``our-2s” means that the output feature map is 1/2 of the input image, and the feature map $ P_1 $ shown in Fig.~\ref{fig:fg3} as the input of the iterative model. There are some abbreviations in the table for typesetting, and ``R”, ``P”, and ``F” represent Recall, Precision, and F-measure, respectively. Unless otherwise illustrated, Scale Expansion Algorithm~\cite{CVPR19_PSENet} (``pse") is used as the region growth algorithm by default.)

\subsection{Ablation Study}\label{Ablation-Study}
\textbf{Influence of $ \alpha $ values and the iterative number ($ n $).} We investigate the effect of different $ \alpha $  values in iteration and explore the relationship between detection performance and iterative number ($ n $). The details of $ \alpha $ selection are described in Section~\ref{Probability_Maps} and Eq.~\ref{alpha_select}. We perform experiments on Total-text with no extra pre-training. In training, we randomly crop text regions, resize them to $ 640 \times 640 $, and train the model with batch size 6 for 400 epochs. The Adam optimizer is adopted with an initial learning rate $ 10^{-4} $ and multiplied by $ 0.9 $ after every 100 epochs. Fig.$ \, $\ref{fig:carve_line} shows some typical experimental results, and we can find that the F-measure keeps rising along with the number of iterations ($ n $) increases and begins to tend to level off when $ n \ge 4$. Besides, we notice that our model needs more iterative steps to achieve good performance if the value interval ($ k $) of $ \alpha $ is small ($ k<=2 $). In addition, when all curves tend to level off, the maximum value of alpha ($ \alpha_{n} $) is about greater than 10. To balance efficiency and performance, we set the iterative number of IM to 4 and set the values of corresponding $ \alpha $ to $ (1, 4, 7, 10) $ in all the following experiments.

\begin{figure*}[htbp]
	\subfigcapskip=5pt
	\centering
	\subfigure[]{
		\begin{minipage}[t]{0.195\linewidth}
			\centering
			\includegraphics[width=3.6cm,height=13cm]{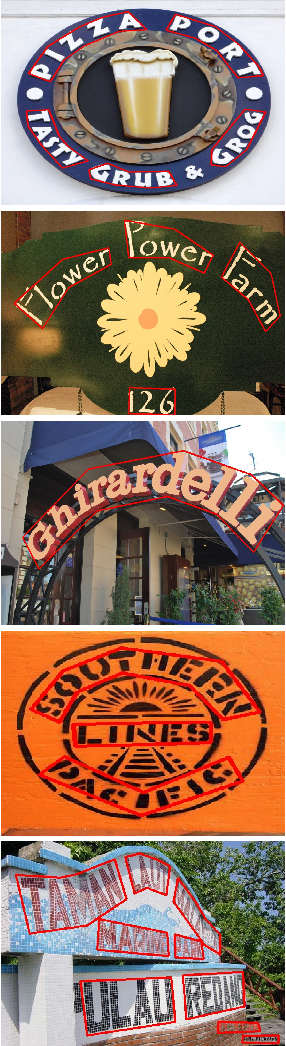}
		\end{minipage}%
	}%
	\subfigure[]{
		\begin{minipage}[t]{0.195\linewidth}
			\centering
			\includegraphics[width=3.6cm,height=13cm]{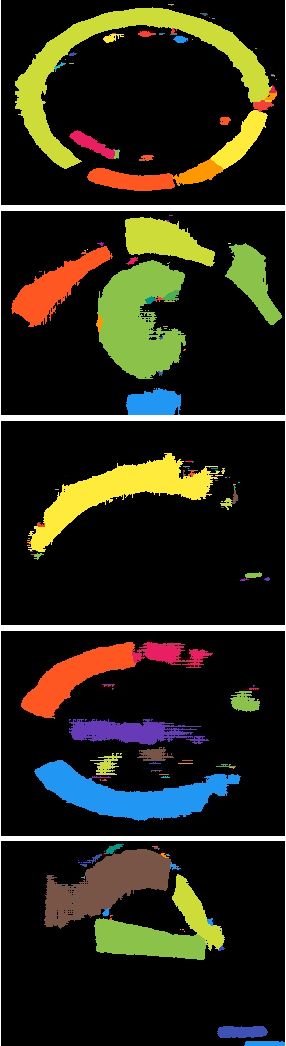}
		\end{minipage}%
	}%
	\subfigure[]{
		\begin{minipage}[t]{0.195\linewidth}
			\centering
			\includegraphics[width=3.6cm,height=13cm]{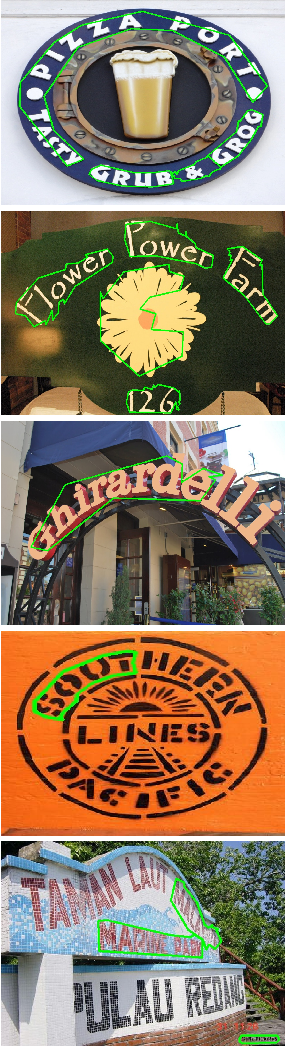}
		\end{minipage}%
	}%
	\subfigure[]{
		\begin{minipage}[t]{0.195\linewidth}
			\centering
			\includegraphics[width=3.6cm,height=13cm]{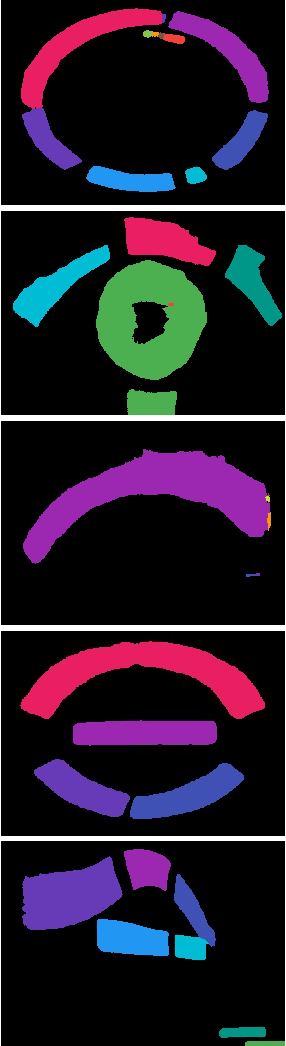}
		\end{minipage}%
	}%
	\subfigure[]{
		\begin{minipage}[t]{0.195\linewidth}
			\centering
			\includegraphics[width=3.6cm,height=13cm]{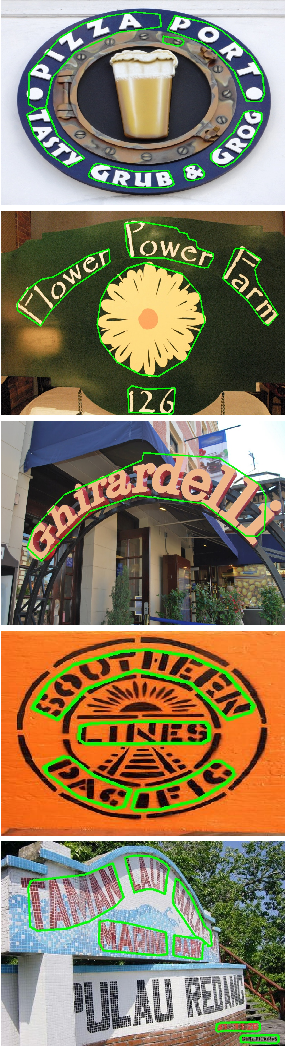}
		\end{minipage}%
	}%
	\centering
	\caption{ Comparisons of segmentation results. (a) Ground-truth annotation boundaries; (b) Segmentation results of text instances supervised by binary maps; (c) The generated contours from (b) via threshold filtering (Algorithm 1); (d) Segmentation results of text instances supervised by probability maps; (e) The generated contours from (d) via Algorithm 1.}
	\label{fig:fig_case_study}
	\vspace{-0.8em}
\end{figure*}

\begin{table}[htbp]
	\begin{center}
		\renewcommand{\arraystretch}{1.3}
		\caption{Ablation study for probability maps (PMS) and iterative module (IM) on Total-Text and MSRA-TD500.}
		\label{table:Ablation}
		\begin{tabular}{c|cccccc}
			\hline
			{\textbf{Datasets}}&{\textbf{PMS}} & {\textbf{IM}} & \textbf{Recall}& \textbf{Precision} & \textbf{F-measure} & \textbf{FPS} \\
			\hline
			\multirow{4}*{Total-Text} 
			&{$ \times $} &{$ \times $}  &80.29 &87.86  &83.82 &5.6\\
			&{$ \times $} &{$ \checkmark $} &80.97 &88.74 &84.68 &5.6\\
			&{$ \checkmark $}  &{$ \times $} &84.33 &87.93  &86.09 &\textbf{6.8}\\
			&{$ \checkmark $} &{$ \checkmark $} &\textbf{84.77} &\textbf{89.81}  &\textbf{87.22} &\textbf{6.8}\\
			\hline
			\hline
			\multirow{4}*{TD500} 
			&{$ \times $} &{$ \times $} &80.58 &81.57 &81.07 &7.4\\
			&{$ \times $} &{$ \checkmark $} &80.24 &83.24  &81.71 &7.4\\
			&{$ \checkmark $}  &{$ \times $} &80.72 &84.31  &82.48 &\textbf{8.3}\\
			&{$ \checkmark $} &{$ \checkmark $} &\textbf{82.65} &\textbf{88.58}  &\textbf{85.51} &\textbf{8.3}\\
			\hline
		\end{tabular}
	\end{center}%
\end{table}

\begin{table*}[htbp]
	\begin{center}
		\renewcommand{\arraystretch}{1.2}
		\caption{Experimental results on Total-Text, CTW-1500, and MSRA-TD500. The symbol $^*$  means a multi-scale test is performed. ``S" means that we use the SynText to pretrain our model, and ``M" means that we use the ICDAR2017-MLT to pretrain our model. ``pse" means we use the Progressive Scale Expansion Algorithm as our post-processing, and ``watershed" means we use the Watershed Algorithm as our post-processing. $ ^{\dagger} $ means voting filtering algorithm is used; otherwise a threshold filtering algorithm is used by default. The best score is highlighted in \textbf{bold}.}	\label{table:tbSyn}
		\begin{tabular}{|c|c||c|c|c|c||c|c|c|c||c|c|c|c|}
			\hline
			\multicolumn{1}{|c|}{ \multirow{2}*{ \textbf{Methods}}}
			&\multicolumn{1}{|c||}{ \multirow{2}*{ \textbf{Published}}}
			& \multicolumn{4}{c|}{\textbf{   Total-Text}} 
			& \multicolumn{4}{c||}{\textbf{ CTW-1500}}
			& \multicolumn{4}{c|}{\textbf{MSRA-TD500}}\\
			\cline{3-14}
			&&\textbf{R}
			& \textbf{P} & \textbf{F}&
			\textbf{FPS}&
			\textbf{R}& 
			\textbf{P} & \textbf{F}&
			\textbf{FPS}&
			\textbf{R}& 
			\textbf{ P} & \textbf{F} &
			\textbf{FPS}\\
			\hline
			SegLink \cite{SegLink}&CVPR2017 &- &- &-&-&- &- &-&- &70.0 &86.0 &77.0&8.9\\
			MCN \cite{MCN}& CVPR2018&- &- &- &-&- &- &-&- &79 &88 &83&-\\
			TextSnake \cite{TextSnake}&ECCV2018&74.5 &82.7 & 78.4&- &\textbf{85.3} &67.9 & 75.6&- &73.9 &83.2 &78.3&1.1\\
			LSE\cite{CVPR19_LSA}&CVPR2019 &- &- &- &- &77.8 &82.7 & 80.1&- &81.7& 84.2 &82.9&-\\
			ATTR\cite{CVPR19_ATRR}&CVPR2019 &76.2 &80.9 & 78.5 &10.0 &- &- &- &- &82.1 &85.2 & 83.6 &-\\
			MSR\cite{MSR}&IJCAI2019 &73.0 &85.2 & 78.6 &- &79.0 &84.1 & 81.5 &- &76.7 &87.4 &81.7&-\\
			CSE\cite{CVPR19_CSE}&CVPR2019 &79.7&81.4&80.2&0.42&76.1&78.7&77.4&0.38&- &- &-&-\\
			TextDragon\cite{TextDragon}&ICCV2019&75.7&85.6&80.3&-&82.8&84.5&83.6&-&- &- &-&-\\
			TextField\cite{TextField}&TIP2019 &79.9&81.2&80.6&-&79.8&83.0&81.4&-&75.9 &87.4 & 81.3&5.2\\
			PSENet-1s \cite{CVPR19_PSENet}&CVPR2019 &77.96&84.02&80.87&3.9&79.7&84.8& 82.2&3.9&- &- &-&-\\
			ICG \cite{SegLink++}&PR2019&80.9&82.1& 81.5&-&79.8&82.8&81.3&-&- &- &-&-\\
			LOMO*\cite{CVPR19_LOMO}&CVPR2019 &79.3&87.6&83.3&-&76.5&85.7&80.8 &-&- &- &-&-\\
			CRAFT \cite{CRAFT}&CVPR2019 &79.9&87.6&83.6&-&81.1&86.0&83.5&-&78.2 & 88.2 &82.9&8.6\\
			DB\cite{DB}&AAAI2020 &82.5 &87.1  &84.7 &{32.0}&80.2 &86.9 &83.4&{22.0}&79.2 &91.5 &84.9&\textbf{32.0}\\
			PAN\cite{PSENet_v2}&ICCV2019 &81.0 &89.3  &85.0&\textbf{39.6} &81.2 &86.4 &83.7&\textbf{39.8}&83.8 &84.4  &84.1&30.2\\
			TextPerception \cite{TextPerception }&AAAI2020 &81.8&88.8&85.2&-&81.9&87.5&84.6&- &-  &- &-&-\\
			ContourNet~\cite{ContourNet}&CVPR2020&83.9&86.9&85.4&3.8&84.1&83.7&83.9 &4.5&-&-&-&-\\
			DRRG~\cite{DRRG}& CVPR2020&84.93&86.54&85.73&- &83.02&85.93&84.45&-&82.30&88.05&85.08&-\\
			Boundary\cite{Boundary}&AAAI2020 &85.0 &88.9  &87.0&- &- &- &- &-&- &-  &- &-\\
			Textmountain\cite{TextMountain}&PR2021 &- &- &- &- &83.4 &82.9 &83.2 &-&- &-  &- &-\\  
			\hline
			\hline
			\textbf{Ours-1s}(S+pse)&- & 84.77 &89.81 &87.22 &6.8 &80.80 &87.60 &84.06 &9.0 &82.65 &88.58& 85.51 &8.3\\
			\textbf{Ours-2s}(M+pse)&- &85.02 &{89.85} &{87.37} &{13.8} &82.14 &86.15&84.10 &14.3 &81.42&88.28&84.71&14.0\\
			\textbf{Ours-1s}(M+pse)&- &\textbf{87.46} &{89.74} &{88.58} &6.8 &83.77 &87.12&85.41 &9.0 &85.91&\textbf{91.40}&88.57&8.3\\
			\textbf{Ours-1s}(M+pse)$ ^{\dagger} $&-&86.11 &\textbf{91.49} &88.72&1.2&83.90 &\textbf{88.39}&\textbf{86.09}&2.5&\textbf{86.08} &90.44&88.21&3.0\\
			\hline
			\textbf{Ours-1s}(M+watershed)&-&87.67 &89.95 &\textbf{88.79}&6.98&83.83 &87.75&85.75&9.14&\textbf{86.94} &91.01&\textbf{88.93}&10.58\\
			\hline
		\end{tabular}
	\end{center}%
	\vspace{-1.5em}
\end{table*}

\noindent \textbf{Influence of probability maps (PMS) and iterative module (IM).} We conduct ablation studies on Total-Text and MSRA-TD500 datasets with pre-training on SynText to verify the effectiveness of the proposed probability maps (PMS) and iterative module (IM). A part of the train details is described in Section ~\ref{Implement-Details}. For Total-Text, we use $ 640 \times 640 $ images to train our model with 300 epochs and $ 800 \times 800 $ images with 300 epochs. For MSRA-TD500, we only use $ 640 \times 640 $ images to train our model with 1,200 epochs. To verify the effectiveness of probability maps (PMS), We replace the probability maps with a group shrieked binary masks similar to~\cite{CVPR19_PSENet}. In this way, we can use the Progressive Scale Expansion Algorithm in~\cite{CVPR19_PSENet} to reconstruct text instances with probability maps or binary masks. For a fair comparison, we adjust the parameters appropriately to get optimum performance. 

As listed in Tab.~\ref{table:Ablation}, we can conclude that the proposed PMS and IM significantly improve the performance on Total-Text (word-level curve texts) and TD500 (line-level long texts). PMS achieves 2.54\% (on Total-Text) and 3.8\% (on TD500) performance improvements in terms of F-measure. IM achieves 1.13\% (on Total-Text) and 3.03\% (on TD500) performance improvement in terms of F-measure. In conclusion, our PMS and IM can enhance performance by 3.4\% (on Total-Text) and 4.44\% (on TD500) in terms of F-measure.

\noindent \textbf{Case study of probability maps.} To further verify the effectiveness of our SAF strategy, we select some real examples in our case study, which are shown in Fig.~\ref{fig:fig_case_study}. In this case study, we respectively train our network with the supervision of binary maps and probability maps for comparison. As shown in Fig.~\ref{fig:fig_case_study} (b), the segmentation results under the supervision of binary maps contain a lot of noises, which may lead to text adhesion or miss detection. As shown in the 1st row of Fig.~\ref{fig:fig_case_study} (c), different text instances are adhesion together. If we filter the noises, some text with low confidence is inevitably eliminated, as shown in the 3rd and 4th row of Fig.~\ref{fig:fig_case_study} (c). With the supervision of our probability maps, the segmentation results are very clean. As shown in Fig.~\ref{fig:fig_case_study} (d), our method can well separate different text instances and reduce the occurrences of missed detection. In addition, because probability maps contain the size information of the text, which can improve the capability of our network for detecting large text, as shown in the 3rd row of Fig.~\ref{fig:fig_case_study} (e). Apparently, the network trained by the supervision of binary maps fails in detecting large text, as shown in the 3rd row of Fig.~\ref{fig:fig_case_study} (c). In addition, with probability maps, our final detected contours are smooth, as shown in the 2nd row of Fig.~\ref{fig:fig_case_study} (e). However, with binary maps, the final detected contours contain a lot of defects and burrs, as shown in the 1st row of Fig.~\ref{fig:fig_case_study} (c).

\subsection{Comparisons with State-of-the-Art Methods}\label{main_result}
Obviously, the predictions accuracy of probability maps will be improved along with iterating. Therefore, when using the voting filtering algorithm, we set different voting weights for each probability map and then selectively filter text instances according to the weighted voting. In the following experiments, the voting weights are fixed at ($ {0.1, 0.2, 0.3, 0.4} $), and the details of voting filtering are referred to  Algorithm \ref{algo:vote}.

\begin{figure*}[htbp]
	\subfigcapskip=-3pt
	\centering
	\subfigure[Total-Text]{
		\begin{minipage}[t]{0.245\linewidth}
			\centering
			\includegraphics[width=4.45cm,height=6cm]{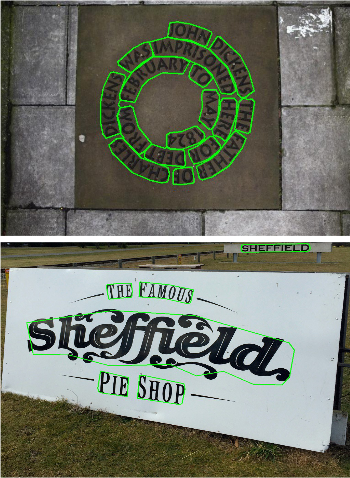}
		\end{minipage}
	}%
	\subfigure[Total-Text]{
		\begin{minipage}[t]{0.245\linewidth}
			\centering
			\includegraphics[width=4.45cm,height=6cm]{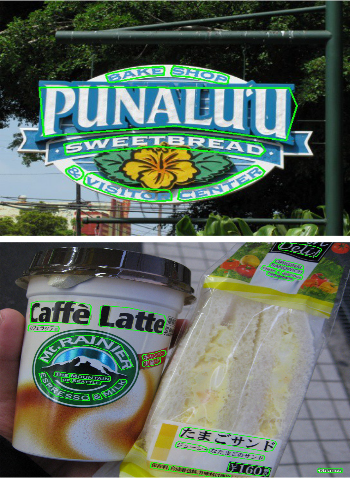}
		\end{minipage}
	}%
	\subfigure[CTW1500]{
		\begin{minipage}[t]{0.245\linewidth}
			\centering
			\includegraphics[width=4.45cm,height=6cm]{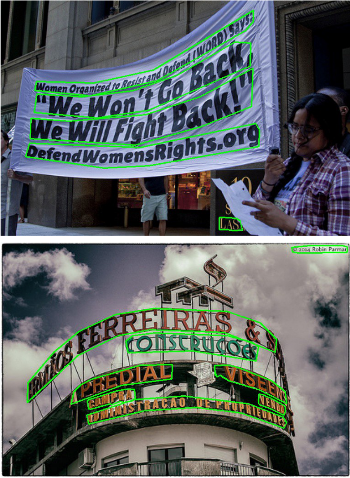}
		\end{minipage}
	}%
	\subfigure[CTW1500]{
		\begin{minipage}[t]{0.245\linewidth}
			\centering
			\includegraphics[width=4.45cm,height=6cm]{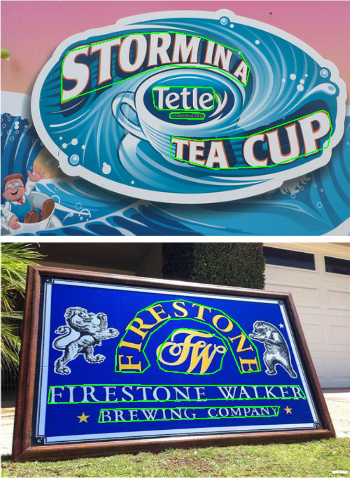}
		\end{minipage}
	}%
	\centering
	\caption{ Some qualitative detection results on Total-Text in (a) and(b), and SCUT-CTW500 in (c) and (d). The arbitrary-shaped texts are correctly detected with accurate text instance boundaries.}
	\label{fig:result_1}
	\vspace{-0.8em}
\end{figure*}

\textbf{Curved text detection}: We restrict the long side of input images not greater than 1,024 and scale the short side of input images to 256 and 512 for Total-Text and CTW-1500 datasets, respectively. In training, we firstly crop $ 640 \times 640 $ image regions to train for 300 epochs and then use $ 800 \times 800 $ image regions to train for 300 epochs. 
During inference, we use the threshold filtering algorithm and the voting filtering algorithm to test, respectively. Some visual results are shown in Fig.~\ref{fig:result_1}, and detailed experimental results are listed in Tab.~\ref{table:tbSyn}.

When threshold filtering is performed, $ th_{b} $ is set to 0.3, and $ th_{e} $ is set to 0.65 on both curve datasets. As listed in Tab.~\ref{table:tbSyn}, our method achieves surprising performance on Total-Text and CTW-1500. Specifically, our method outperforms PAN by 3.58\%, DB by 3.88\%, and  CRAFT by 4.98\% in terms of F-measure on Total-Text. And our method also outperforms PAN by 1.71\%, DB by 2.01\%, and CRAFT by 1.91\% in terms of F-measure on CTW-1500. 

When voting filtering is performed, $ th_{b} $ is set to 0.325 for Total-Text and 0.365 for CTW-1500. 
The results listed in Tab.~\ref{table:tbSyn} show that the voting filtering algorithm not only reduces the threshold parameters meanwhile increases the adaptability of our model but also slightly improves the detection performance of our model (0.14\% on Total-Text, 0.68\% on CTW-1500). Notably, our method achieves the best performance in curve text detection (88.72\% on Total-Text, 86.09\% on CTW-1500 in terms of F-measure).

\begin{table}[htbp]
	\begin{center}
		\renewcommand{\arraystretch}{1.2}
		\caption{Experimental results on ICDAR17-MLT.}
		\label{table:MLT}
		\begin{tabular}{c||ccc}
			\hline
			\textbf{Methods}& \textbf{Recall}& \textbf{Precision} & \textbf{F-measure}\\
			\hline
			Ma et al. \cite{RRPN} &55.50 &71.17 &62.37\\ 
			He et al. \cite{MOML} &57.9 &76.7  &66.0\\ 
			Border \cite{ASTD} &60.6 &73.9  &66.6\\ 
			Corner.\cite{corner} &55.6 &\textbf{83.8} &66.8\\
			FOTS\cite{FOTS} &57.51 &80.95 &67.25\\
			DRRG~\cite{DRRG}&61.04&74.99&67.31\\
			LOMO\cite{CVPR19_LOMO} &60.6 &78.8 &68.5\\
			PSENet-1s \cite{CVPR19_PSENet}&\textbf{68.21}&73.77&70.88\\
			DB-ResNet-18\cite{DB}&63.8&81.9 &71.7\\
			DB-ResNet-50\cite{DB}&67.9 &83.1 &74.7\\
			\hline
			\hline
			\textbf{Ours-1s}(pse)&64.81 &81.24 &72.10\\
			\textbf{Ours-1s}(pse)$ ^{\dagger} $ &65.94 &80.56 &\textbf{72.52}\\
			\hline
			\textbf{Ours-1s}(watershed) &65.15&81.72&72.50\\
			\hline
		\end{tabular}
	\end{center}%
	\vspace{-1.2em}
\end{table}

\begin{table}[htbp]
	\begin{center}
		\renewcommand{\arraystretch}{1.1}
		\caption{Experimental results on ICDAR2015. The symbol $^*$  means Multi-scale test is performed.}
		\label{table:ICDAR15}
		\begin{tabular}{c||ccc}
			\hline
			\textbf{Methods}& \textbf{Recall}& \textbf{Precision} & \textbf{F-measure}\\
			\hline
			SegLink \cite{SegLink} &76.8 &73.1 &75.0\\
			MCN \cite{MCN} &72 &80 &76\\
			RRPN$^*$ \cite{RRPN} &77 &84 &80\\
			EAST$^*$ \cite{EAST} &78.3 &83.3 &80.7\\
			He et al. \cite{DDR_TIP} &80 &85 &82\\
			RRD \cite{RRD} &79 &85.6 & 82.2\\
			TextField \cite{TextField} &80.05 &84.3 &82.4\\
			TextSnake \cite{TextSnake} &84.9 &80.4 &82.6\\
			Textboxes++$^*$\cite{textboxes++} &78.5 &87.8 &82.9\\
			PAN\cite{PSENet_v2} &84.0 &81.9  &82.9\\
			PixelLink \cite{PixelLink} &82.0 &85.5 &83.7\\
			FTSN \cite{FTSN}  &80.0 &88.6 &84.1\\
			FOTS \cite{FOTS}&82.04 &88.84 &85.31\\
			PSENet-1s\cite{CVPR19_PSENet} &84.5 &86.92 & 85.69\\
			Textmountain\cite{TextMountain}&84.16 &88.51 &86.28\\
			DRRG~\cite{DRRG}&84.69&88.53 &86.56\\
			LSE \cite{CVPR19_LSA} &\textbf{85.0} &88.3 &86.6 \\
			ATRR \cite{CVPR19_ATRR} &83.3 &90.4 &86.8\\
			CRAFT \cite{CRAFT} &84.3 &89.8 &86.9 \\
			ContourNet~\cite{ContourNet}&86.1 &87.6&86.9\\
			LOMO \cite{CVPR19_LOMO} &83.5 &91.3 &87.2 \\
			DB-ResNet-18\cite{DB}&78.4&86.8  &82.3\\
			DB-ResNet-50\cite{DB}&83.2&\textbf{91.8} &87.3\\
			\hline
			\hline
			\textbf{Ours-1s} (S+pse) &83.29&88.18&85.66\\
			\textbf{Ours-1s} (M+pse) &84.50&90.18&87.25\\
			\textbf{Ours-1s} (M+pse)$ ^{\dagger} $ &84.74&89.02&86.83\\
			\hline
			\textbf{Ours-1s}(M+watershed) &84.93&89.91&\textbf{87.35}\\
			\hline
		\end{tabular}
	\end{center}%
	\vspace{-1.2em}
\end{table}

\textbf{Multi-language text detection}: To verify the effectiveness of our method, we conduct experiments on two multi-language datasets. For MSRA-TD500, we restrict the long side of input images to 832, but there is no restricted size of the short side. In training MSRA-TD500, we use $ 640 \times 640 $ image regions to train for 1,200 epochs. For threshold filtering, $ th_{b} $ and $ th_{e} $ are set to 0.305 and 0.8, respectively. For voting filtering $ th_{b} $ is also set 0.305. As listed in Tab.~\ref{table:tbSyn}, our method achieves state-of-the-art performance on MSRA-TD500. Although the voting filtering algorithm will cause a bit of performance loss, the detection performance of our method is still amazing. On MSRA-TD500, our method outperforms PAN by 4.47\%, DB by 3.67\%, and CRAFT by 5.67\% in terms of F-measure with threshold filtering.

For the ICDAR2017-MLT dataset, we restrict the long side of input images no greater than 1920 and scale the short edge of the input image to 256. In training, we firstly crop $ 640 \times 640 $ image regions to train for 150 epochs and then use $ 800 \times 800 $ image regions to train for 150 epochs.  For threshold filtering, $ th_{b} $ and $ th_{e} $ are set to 0.405 and 0.8, respectively. For voting filtering, $ th_{b} $ is also set 0.345. As listed in Tab.~\ref{table:MLT}, our method also achieves impressive performance against other methods in detecting complex multi-language and multi-oriented texts. On ICDAR2017-MLT,  our method outperforms PSENet-1s by 1.64\% and LOMO by 4.02\% in terms of F-measure.

\begin{figure*}[htbp]
	\subfigcapskip=-3pt
	\centering
	\subfigure[MSRA-TD500]{
		\begin{minipage}[t]{0.245\linewidth}
			\centering
			\includegraphics[width=4.45cm,height=6cm]{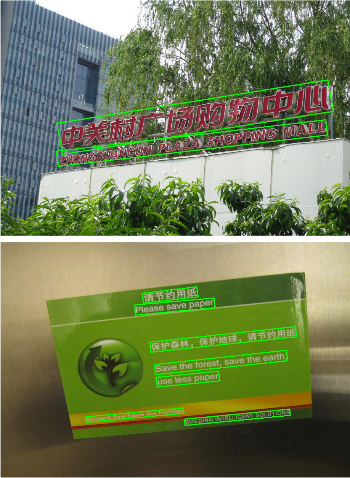}
		\end{minipage}
	}%
	\subfigure[MSRA-TD500]{
		\begin{minipage}[t]{0.245\linewidth}
			\centering
			\includegraphics[width=4.45cm,height=6cm]{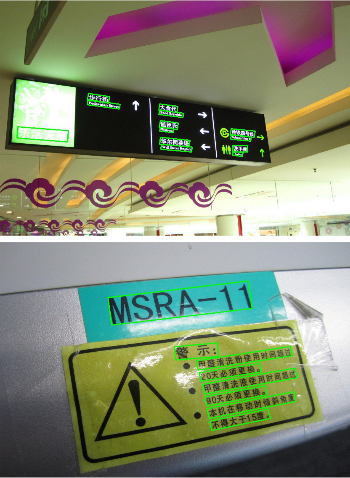}
		\end{minipage}
	}%
	\subfigure[ICDAR2017-MLT]{
		\begin{minipage}[t]{0.245\linewidth}
			\centering
			\includegraphics[width=4.45cm,height=6cm]{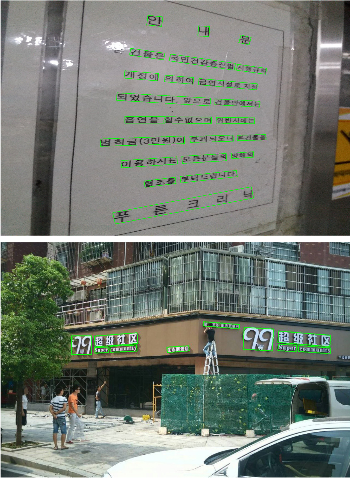}
		\end{minipage}
	}%
	\subfigure[ICDAR2015]{
		\begin{minipage}[t]{0.245\linewidth}
			\centering
			\includegraphics[width=4.45cm,height=6cm]{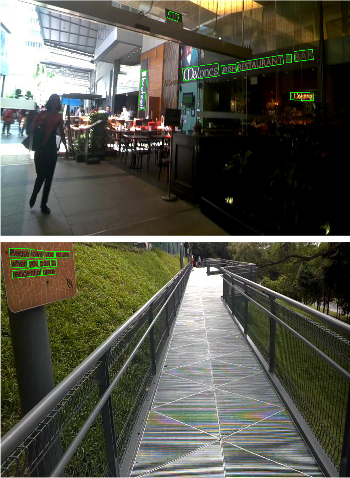}
		\end{minipage}
	}%
	\centering
	\caption{Representative qualitative detection results on MSRA-TD500 in (a) and (b), ICDAR2015 in (c), and ICDAR2017-MLT in (d). The multi-oriented and multi-lingual texts are correctly detected with accurate text instance boundaries. Both line level-text detection and word-level text detection can accurately detect the boundaries of text instances.}
	\label{fig:result_2}
	\vspace{-0.8em}
\end{figure*}

Some qualitative illustrations are shown in Fig.~\ref{fig:result_2}~(a) and (b) for MSRA-TD500, and Fig.~\ref{fig:result_2}~(c) for ICDAR2017-MLT. Our proposed method successfully detects long text lines of arbitrary orientations and sizes, and also successfully detects complex multi-language texts.

\begin{table}[htbp]
	\begin{center}
		\renewcommand{\arraystretch}{1.0}
		\caption{ Analysis of time consumption. Time consists of the backbone (B), iterative module (IM), Data transfer(T), and post-processing (Post). “Res” denotes the resolution of the input image. Data transfer refers to the time of transferring output CPU from GPU.}	\label{table:speed}
		\begin{tabular}{|c|c|c||c|c|c|c||c|}
			\hline
			\multicolumn{1}{|c|}{ \multirow{2}*{\textbf{Methods}}}
			&\multicolumn{1}{|c|}{ \multirow{2}*{\textbf{Res}}}
			&\multicolumn{1}{c||}{ \multirow{2}*{\textbf{F}}}
			&\multicolumn{4}{c||}{\textbf{Time consumption (ms)}}
			& \multicolumn{1}{|c|}{ \multirow{2}*{\textbf{FPS}}}\\
			\cline{4-7}
			&&&\textbf{B}
			& \textbf{IM} 
			& \textbf{T}
			& \textbf{Post}&\\
			\hline
			\hline
			\multicolumn{8}{|c|}{\textbf{Total-Text}}\\
			\hline
			ours-1s&1024&88.58&24&1.1&47 &75&6.8\\
			ours-2s&1024&87.37&22&1.0&32&17&13.8\\
			ours-2s&800&85.69&22&1.0&20&12&17.6\\
			ours-2s&640&83.97&21&0.9&14&8&22.5\\
			\hline
			\hline
			\multicolumn{8}{|c|}{\textbf{CTW-1500}}\\
			\hline
			ours-1s&1024&85.41&17&0.9&45&49&9.0\\
			ours-2s&1024&84.10&16&0.9&33&20&14.3\\
			ours-2s&800&83.38&16&0.9&25&13&18.3\\
			ours-2s&640&82.67&16&0.8&14&9&25.2\\
			\hline
		\end{tabular}
	\end{center}%
	\vspace{-1.5em}
\end{table}

\textbf{Multi-oriented text detection}: The ICDAR2015 is a typical multi-oriented text dataset that contains lots of small and
low-resolution text instances. Due to the images in the ICDAR2015 testing dataset having an identical size, we scale the short side of input images to 960. In training, we use $ 640 \times 640 $ image regions to train for 400 epochs, and then use $ 800 \times 800 $ image regions to train for 600 epochs. When pre-training on ICDAR2017-MLT,  the training images in ICDAR2015, the training and testing images in ICDAR2013 are used to fine-tune our network.  

In testing, $ th_{b} $ and $ th_{e} $ are set to 0.405 and 0.84 for threshold filtering, respectively. For voting filtering, $ th_{b} $ is set to 0.335. As listed in Tab.~\ref{table:ICDAR15}, our method achieves competitive performance against other methods in detecting small multi-oriented texts. Our method outperforms PSENet by 1.56\% and LSE by 0.65\% in terms of F-measure with threshold filtering. Although the voting filtering results have a minor performance degradation of 0.42\% in F-measure, our performance is still promising. In addition, the voting filtering algorithm reduces the dependence on empirically pre-defined threshold parameters and thus improves the adaptability and robustness of our method. Representative qualitative illustrations are shown in Fig.~\ref{fig:result_2}~(d).

\begin{figure}[htbp]
	\subfigcapskip=-3pt
	\centering
	\subfigure[]{
		\begin{minipage}[t]{0.49\linewidth}
			\includegraphics[width=4.35cm,height=3cm]{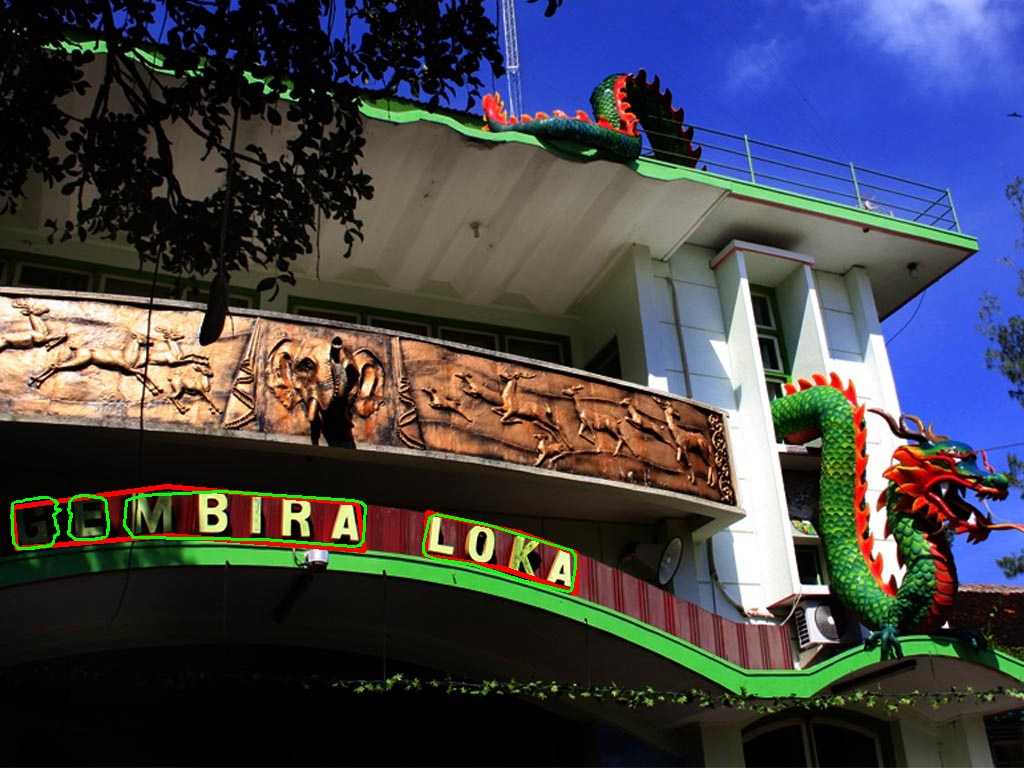}
		\end{minipage}
	}%
	\subfigure[]{
		\begin{minipage}[t]{0.49\linewidth}
			\includegraphics[width=4.35cm,height=3cm]{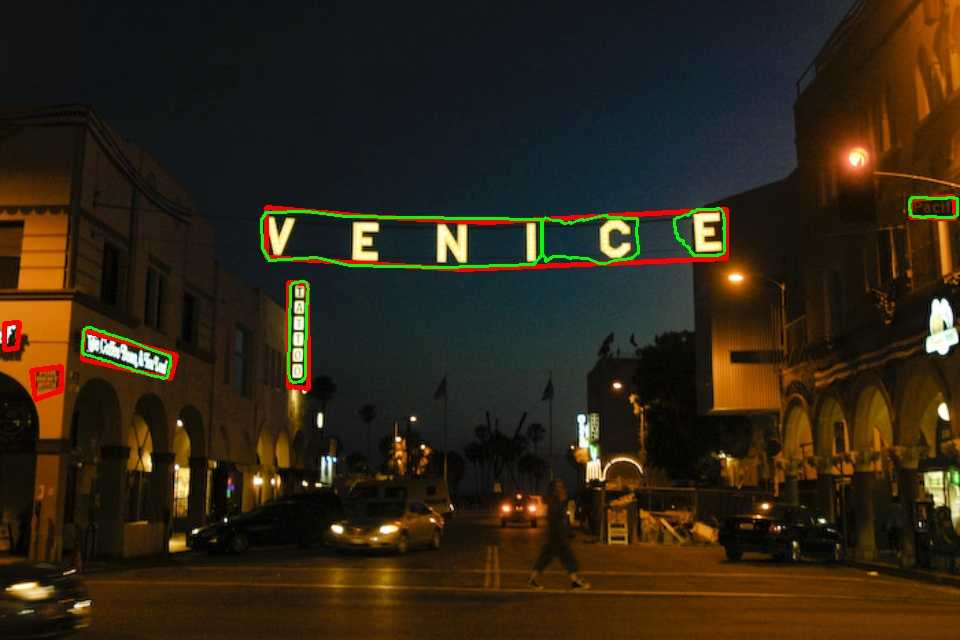}
		\end{minipage}
	}%
	\vspace{-0.6em}
	\\
	\subfigure[]{
		\begin{minipage}[t]{0.49\linewidth}
			\includegraphics[width=4.35cm,height=3cm]{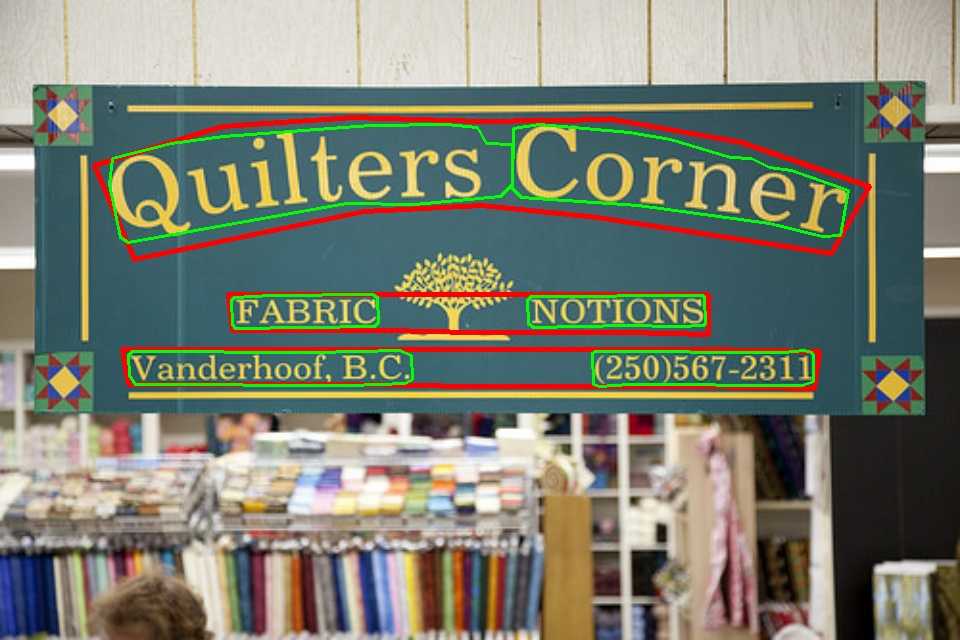}
		\end{minipage}
	}%
	\subfigure[]{
		\begin{minipage}[t]{0.49\linewidth}
			\includegraphics[width=4.35cm,height=3cm]{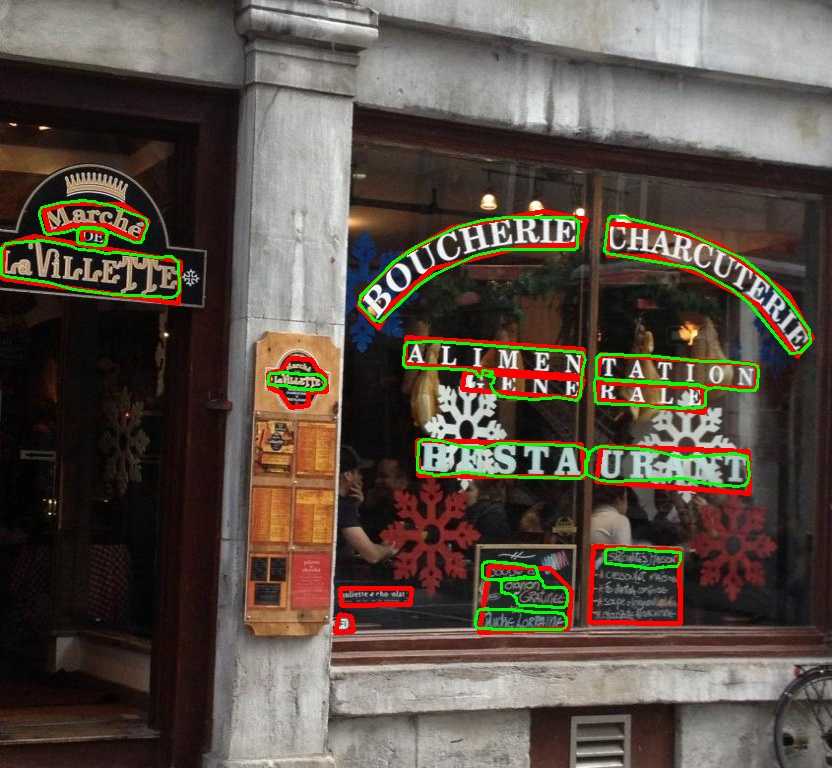}
		\end{minipage}
	}%
	\vspace{-0.6em}
	\\
	\subfigure[]{
		\begin{minipage}[t]{0.49\linewidth}
			\includegraphics[width=4.35cm,height=3cm]{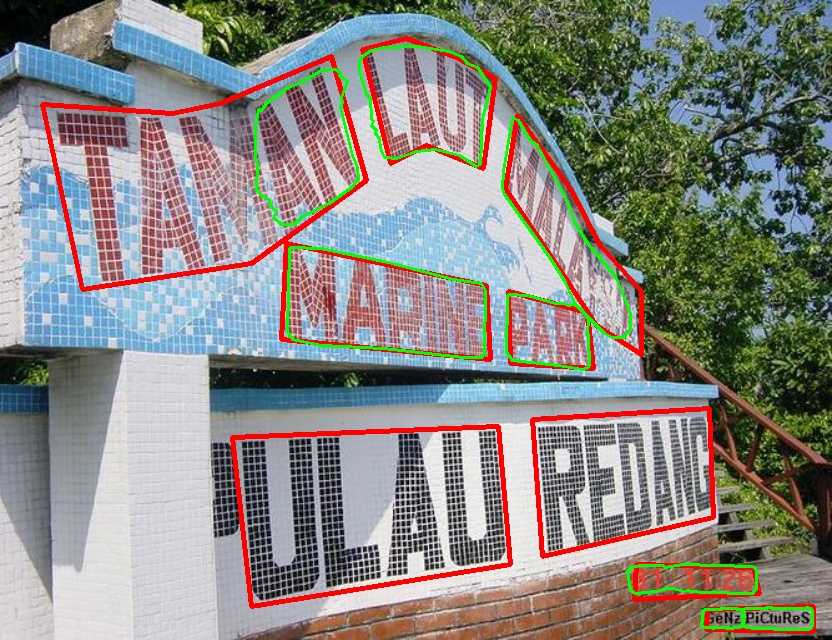}
		\end{minipage}
	}%
	\subfigure[]{
		\begin{minipage}[t]{0.49\linewidth}
			\includegraphics[width=4.35cm,height=3cm]{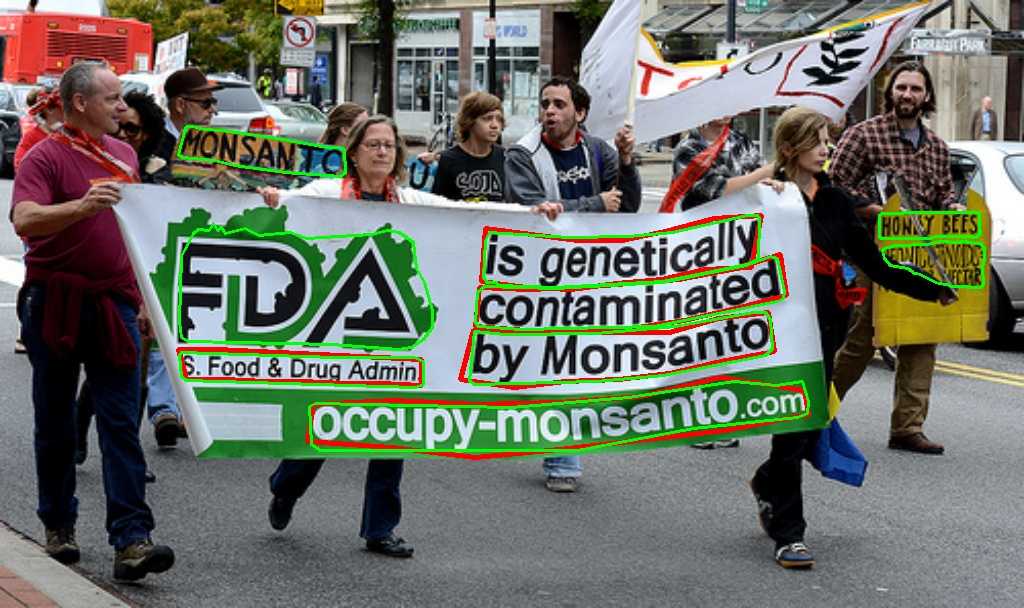}
		\end{minipage}
	}%
	\centering
	\caption{ Visualizations of detection errors. Green contours: detections of our methods with watershed algorithm; Red contours: ground truths.}
	\label{fig:weakness}
\end{figure}

\subsection{Speed Analysis} 
As shown in Table 6, our method can efficiently detect curve text instances. We especially analyze the time consumption of our method in different stages on two curve text datasets. When the output feature map is 1/1 of the input image, the proposed method obtains the best performance, while the time consumption of post-processing is more than half of the total inference time because of the large feature map. If the size of the output feature map is 1/2 of the input images, the FPS of our method boosts from 6.8 to 13.8, while the performance slightly decreases from 88.58\% to 87.37\% on Total-Text, which is shown in Tab.~\ref{table:speed} and Tab.~\ref{table:tbSyn}. On CTW-1500, if the size of the output feature map is 1/2 of the input images, our FPS boosts from 9.0 to 13.8, while the performance slightly decreases from 85.41\% to 84.10\%. Furthermore, when we scale the long edge of 640, our FPS increases to 22.5, and the detector still has good performance (83.97\%), which is better than CRAFT~\cite{CRAFT} on Total-text. The speed of our method is nearly real-time (25.2 FPS), while the performance is still competitive on CTW-1500 when the long edges of input images are restricted to 640.  As listed in Tab.~\ref{table:speed}, the post-processing and data transfer is time-consuming because the region growth algorithms in post-processing is a serial algorithm that executes pixel by pixel processing on the CPU. Therefore, an efficient and parallel region growth algorithm will significantly improve the detection speed of our method.

\subsection{Weakness}
As demonstrated in the experiments, our method achieves superior performance in detecting texts of arbitrary shapes. However, there are still some failing cases, especially on challenging images, such as images with object occlusion, large character spacing, or text-like areas. Some representative failure examples are given in Fig.~\ref{fig:weakness}. As shown in Fig.~\ref{fig:weakness} (a-b), our method fails in covering the whole text instance because of the large character spacing. This is caused by the not wholly covered boundary proposal for the text due to the limited receptive field. Fig. 13 (c-d) shows that some object occlusions and text-like areas lead to poor detections. As shown in Fig.~\ref{fig:weakness} (e), limited by the receptive field and imbalance training samples, some extremely large text is still challenging for our method. But, all these difficulties are common challenges for the other methods~\cite{DB, PSENet_v2, TextField, CVPR19_PSENet}. Moreover, there are some ``false detections” caused by unreasonable or missing annotations, as shown in Fig.~\ref{fig:weakness} (c) and (f).

\section{Conclusion}\label{Conclusion}
This paper presents a novel and robust segmentation-based text detector for arbitrary shape text detection, employing multiple probability maps to segment the text instance. We adopt a group of probability maps with a series of Sigmoid Alpha Functions ($ SAF $) to describe the possible probability distributions as much as possible. To predict high-precision probability maps, we propose an iterative model to optimize the predictions of probability maps by implicitly learning the mapping relationships between successive probability distributions. A variety of experiments have verified that our method consistently outperforms state-of-the-art methods in terms of detection accuracy on a variety of typical scene text benchmarks. In the future, we would explore a more efficient text reconstruction algorithm to improve our method's detection speed further and design an efficient end-to-end text reading system.

\ifCLASSOPTIONcaptionsoff
  \newpage
\fi

\bibliographystyle{IEEEtran}
\bibliography{Reference}

\end{document}